%% file: neurips_2023.tex
\documentclass{article}

\usepackage[citestyle=numeric,bibstyle=numeric,maxcitenames=2,maxbibnames=9,uniquelist=false]{biblatex}
\usepackage[nonatbib, preprint]{neurips_2023} 

\usepackage{microtype}
\usepackage{graphicx}
\usepackage{subcaption}
\usepackage{booktabs} 

\usepackage{hyperref}

\usepackage{algorithm}
\usepackage{algorithmic}

\usepackage{amsmath}
\usepackage{amssymb}
\usepackage{mathtools}
\usepackage{amsthm}
\usepackage{enumitem}

\usepackage[capitalize,noabbrev]{cleveref}
\usepackage{multirow}
\theoremstyle{plain}

\theoremstyle{definition}

\theoremstyle{remark}

\newcommand{\cE}{\mathcal{E}}

\newcommand{\cN}{\mathcal{N}}
\newcommand{\cP}{\mathcal{P}}

\newcommand{\cR}{\mathcal{R}}
\newcommand{\cS}{\mathcal{S}}
\newcommand{\cT}{\mathcal{T}}

\newcommand{\CLS}{\texttt{CLS}{}}
\newcommand{\R}{\mathbb{R}}

\newcommand{\EE}[2][]{\mathbb{E}_{#1}\left[#2\right]}

\newcommand{\bx}{\mathbf{x}}

\newcommand{\ttbf}[1]{\large{\texttt{\textbf{#1}}}}
\newcommand{\ttbfs}[1]{\texttt{\textbf{#1}}}
\usepackage{amsmath}

\DeclareMathOperator*{\argmin}{arg\,min}
\newcommand{\proj}{\mathrm{proj}}

\newcommand{\restr}{\mathrm{allow}}

\newcommand{\simiid}{\overset{\mathrm{iid}}{\sim} }

\newcommand{\citet}[1]{\textcite{#1}}
\newcommand{\citep}[1]{\parencite{#1}}
\usepackage[most]{tcolorbox}

\newcommand{\tbox}[2][]{
\begin{tcolorbox}[
    colback=gray!20,
    colframe=gray!50,
    coltext=black,
    coltitle=black,
    title={#1},
    left=2pt
]
#2
\end{tcolorbox}
}
\usepackage[textsize=tiny]{todonotes}
\usepackage{CJKutf8}
\usepackage[utf8]{inputenc}

\title{Black Box Adversarial Prompting \\ for Foundation Models}
\author{
    Natalie Maus\footnotemark[1] \\
    {\normalsize \textsc{nmaus@seas.upenn.edu}} \\
    University of Pennsylvania
    \And
    Patrick Chao\thanks{Equal contribution.}\\
    {\normalsize \textsc{pchao@wharton.upenn.edu}} \\
    University of Pennsylvania
    \And
    Eric Wong \\
    {\normalsize \textsc{exwong@seas.upenn.edu }} \\
    University of Pennsylvania
    \And 
    Jacob Gardner \\
    {\normalsize \textsc{jacobrg@seas.upenn.edu}} \\
    University of Pennsylvania
}
\date{}

\begin{document}

\maketitle

\begin{abstract}
\input{abstract}
\end{abstract}

\section{Introduction}
\label{sec: intro}
\input{intro}

\section{Adversarial Prompt Optimization}
\label{sec: prompt opt}
\input{prompt_opt}

\section{Losses and Threat Models}
\label{sec: loss and threats}
\input{loss_threats.tex}

\section{Experiments}
\label{sec: experiments}
\input{experiments}

\section{Background and Related Work}
\label{sec: related work}
\input{related_work}

\section{Conclusion}
\label{sec: conclusion}
\input{conclusion}

\printbibliography

\newpage
\appendix
\onecolumn
\include{appendix}

\end{document}

%% file: abstract.tex
Prompting interfaces allow users to quickly adjust the output of generative models in both vision and language. However, small changes and design choices in the prompt can lead to significant differences in the output. In this work, we develop a black-box framework for generating adversarial prompts for unstructured image and text generation. These prompts, which can be standalone or prepended to benign prompts, induce specific behaviors into the generative process, such as generating images of a particular object or generating high perplexity text.

%% file: intro.tex
Foundation models (FMs) have demonstrated state-of-the-art performance across a diverse range of prediction and generation tasks. Large Language Models (LLMs), such as  GPT-3 \citep{gpt3}, PaLM \citep{palm}, and ChatGPT are capable of answering complex queries when given a short language prompt as instructions. In vision, FMs now come with an easy-to-use text interface that has revolutionized image generation. Text-To-Image Models (TTIM) such as DALL-E 2 \citep{dalle2}, Imagen \citep{imagen}, and Stable Diffusion \citep{StableDiff} are capable of generating a wide range of high-resolution images from text descriptions. 

The common thread amongst all these models is the use of a natural language prompt to influence the output of the model. By adjusting the words in a prompt, users can quickly and easily tweak the FM to generate text or images that better suit their task. This prompting interface has introduced a new mode of interaction with a model---instead of collecting datasets and fine-tuning models for a particular task, one simply changes words in a prompt. 

\begin{figure*}[!ht]
    \begin{subfigure}{.5\textwidth}
    
    \centering
    \begin{center}
\includegraphics[width=0.49\columnwidth]{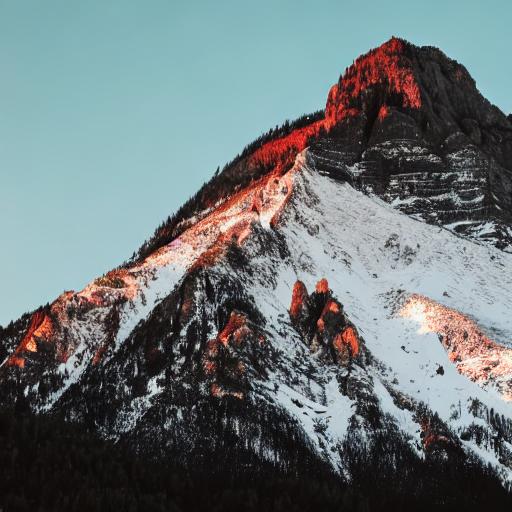}
\includegraphics[width=0.49\columnwidth]{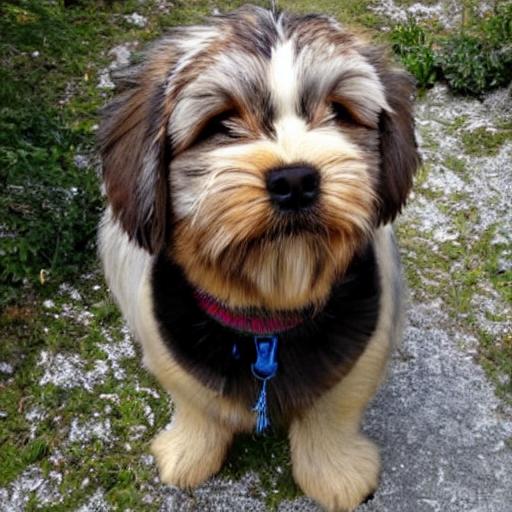}
\caption{Images generated by Stable Diffusion v1.5 \citep{StableDiff} using the prompts \textbf{(Left)} `\texttt{a picture of a mountain}' and \textbf{(Right)} `\texttt{turbo lhaff\checkmark a picture of a mountain}', found by our method. The prepended text causes the model to consistently generate dogs rather than mountains.}
\label{first-page-img}
\end{center}
    \end{subfigure}
    \hspace{0.1cm}
    \begin{subfigure}{.5\textwidth}
        \centering
    \begin{CJK*}{UTF8}{gbsn}
\centering
 \tbox[\textbf{Prompt: }\textit{coordinate hip arts … filled via} Explain list comprehension in Python.]{
            \textbf{Generated Examples: } \\
            \begin{tabular}{l p{5.5cm}}
                 1. &  1. 单词:\textbackslash n 2. 词组: \textbackslash n 3. 短语：\textbackslash n \textbackslash n 关键词：\textbackslash n 1. 单词\\
                 2. & 1. Forward and backward projections are techniques used in machine learning to estimate the gradient of a loss function. \textbackslash 2. In physics, the Lorentz transformation is used to describe how space and time are affected by relative motion.\\ 
                 3. & I'd like to be called ``Alexa," as it is short, simple, and easy to remember. Plus, it has a nice ring to it!
            \end{tabular}
        }
    \caption{Three examples of text generated from Vicuna 13B-v1.1 using our optimized prompt in italics and the original prompt to which we prepend. The generated text are nonsensical or irrelevant to the prompt.}
\label{fig: text-examples}
\end{CJK*}
    \end{subfigure}
    \caption{Examples of adversarial prompts and generated outputs.}
\vspace{-3ex}
\end{figure*}

However, it is becoming apparent that small variations in the prompt can lead to substantial differences in the model output. For example, prompting is known to be sensitive to the order of sentences \cite{lu2021fantastically}, variations in the template \cite{jiang2020can}, and the specific examples provided in the prompt \cite{liu2021makes}. Certain prompts can also result in strange or unusual behavior. For example, gibberish words from generated images such as ``Apoploe vesrreaitais'' can be used as a prompt to generate animals such as birds \cite{daras2022discovering}. 

Recent work has seen researchers crafting adversarial prompts that exploit these vulnerabilities to extract information or induce malevolent behaviors. For example, \citet{perez2022ignore} designed adversarial prompts that induce specific behaviors in an LLM, such as repeating its original prompt or outputting a particular string. \citet{xu2022exploring} studied backdoors and adversarial triggers for LLMs fine-tuned on target tasks. However, these latter adversarial attacks require gradient information and the ability to modify the weights of the LLM. Such attacks are not possible on closed-source, general purpose FMs such as ChatGPT or DALL-E 2 that are used for a wide variety of unstructured tasks. \emph{How do we craft adversarial prompts for general tasks on foundation models with only query access?}

\paragraph{Our Contributions.}
In this paper, we develop a framework for finding adversarial prompts in the black-box setting for general language and vision objectives. We focus specifically on \emph{black-box} attacks as many foundation models are not publicly available, e.g. ChatGPT and DALL-E 2 only have API access. We build upon tools for black-box adversarial attacks traditionally used in the vision setting but instead leverage these tools to search for adversarial prompts. The main challenges are the high dimensionality and discrete nature of the token space. These properties drastically increase the sample complexity and slow down the convergence of classic black-box adversarial attacks. To address these issues, we develop a Token Space Projection (TSP) to map a continuous, lower-dimensional embedding space to the set of discrete language tokens. Our projected variations of black box attacks are capable of finding adversarial prompts, which we demonstrate across a range of vision and language tasks. Our main contributions are as follows: 

\begin{enumerate}
    \item We develop a framework for finding adversarial prompts using a Token Space Projection operator. This operator bridges the continuous word embedding space with the discrete token space, and enables the use of black-box attacks to find adversarial prompts. 
    \item We demonstrate how our framework can automatically find standalone or prepended prompts that cause text-to-image models to output specific image classes. We can further find adversarial prompts that exclude tokens related to the target class. 
    \item Our framework can also find adversarial prompts that alter text generation. For example, we find adversarial prompts that generate irrelevant or incoherent text, or text in different languages.
\end{enumerate}

%% file: prompt_opt.tex
Adversarial attacks in language models have classically focused on perturbing text inputs to change the model's prediction while maintaining the original semantic meaning. Various attack strategies include inducing typos \cite{sun2020adv}, substituting words \cite{jia2019certified}, and paraphrasing \cite{iyyer2018adversarial} to perturb the original text. However, these attacks are designed for a classification setting: they require a clean text example which is then modified. In contrast, there is no such clean example for a prompt, which is designed from scratch. Furthermore, prompt-based models do not generate clean classifications and instead produce comparatively unstructured outputs. 

Therefore, we seek an alternative type of adversarial attack for prompting, one that can induce a targeted change in the output of the generative model. 
As an overview, our adversarial prompt optimization pipeline comprises four steps: 1) generate a candidate word embedding, 2) project the embeddings to a prompt, 3) generate using the prompt to obtain a loss, and 4) use a gradient-free optimization technique to optimize the loss and propose a new candidate word embedding, (\cref{fig: adv prompt pipeline}).

We begin by defining the concept of an \emph{adversarial prompt}.  
Given a prompt $p\in \cP$, the model $M_p$ is represented as a probability distribution over responses $r\in\cR$, parameterized by $p$. For text input models, the arbitrary space $\cP$ is defined  by sequences of tokens with a maximal length $d$. We denote this as $\cP=\cT^d$, with $\cT$ denoting the discrete token space. Let $R_1^p,\ldots,R_n^p\simiid M_p$ be sampled responses from the probability distribution induced by model $M_p$.

Let $\ell:\cR\to \R$ be some external predefined loss function and $\cT_\restr\subseteq \cT$ be a (potentially restricted) subclass of allowed tokens and $\cP_\restr\coloneqq \cT_\restr^d \subseteq \cP$ be a (potentially restricted) subclass of prompts comprised of $d$ tokens from $\cT_\restr$. Consequently, our adversarial prompt framework aims to discover a prompt $p$ that minimizes loss over the distribution of generated responses,
\begin{align}\label{eq: full opt prob}
\argmin_{p\in \cP_\restr} \EE[R\sim M_p]{\ell(R)}.
\end{align}
We relax the optimization problem to an empirical risk minimization framework (ERM) where we observe responses $R_1^{p},\ldots,R_n^{p}\simiid M_{p}$, 
\vspace{-2ex}
\begin{align}\label{eq:erm_loss}
\argmin_{p\in \cP_\restr} \frac{1}{n}\sum_{i=1}^n \ell(R_i^{p}).    
\end{align}
We provide details on using black box optimizers to solve the relaxed ERM formulation in \cref{sec: opt approaches}.

\subsection{Token Space Projection}\label{sec: token space projection}
In our setting, we must optimize over a combinatorially large and discrete input token space $\cP=\cT^d$. To combat this, we consider a word embedding associated with each token and instead optimize over a relaxation of \autoref{eq:erm_loss} defined over the continuous embedding space. 

Consider an embedding $g:\cT\to \cE\subset \R^m$, where for each token $t\in \cT$, we have a unique associated embedding vector $g(t)\in \R^{m}$, where $m$ is the embedding dimension. Any prompt $p=(t_1,\ldots,t_d) \in\cP$ comprised of $d$ tokens can be represented as the embedding $g(p)\coloneqq (g(t_1),\ldots,g(t_d))\in \cE^d\subset \R^{m\times d}$. We note that the embedding $g$ does not need to be the same embedding in the generative model, and we may utilize smaller embeddings for ease in optimization. 

We could immediately optimize over the embedding space $\cE^d$ instead of $\cP$ and obtain adversarial examples, however, we assume only access to a black box generative model with text input functionality and thus are only interested in usable text prompts. To ensure we find text prompts, we project the candidate embeddings $e$ to tokens, and evaluate the loss function with the projected tokens. The projection function $\proj_S: \cE \to \cS$ chooses the closest token in Euclidean distance in a set $\cS$,
\begin{align}
\proj_S(e)\coloneqq\argmin_{t \in \cS} \|g(t)-e\|_2.
\vspace{-1ex}
\end{align}
Denote by $\mathbf{e} = (e_1,\ldots,e_d)$ the embedding matrix corresponding to the $d$ tokens. With slight abuse of notation, let $\proj_{\cS}(\mathbf{e})=(\proj_{\cS}(e_1),\ldots,\proj_{\cS}(e_d))$, mapping embedding matrices to prompts. In our setting, we consider $\proj_{\cT_\restr}(e_i)$, projecting embeddings to the closest tokens in the restricted subset $\cT_\restr$ and, using the above abuse of notation, its prompt extension $\proj_{\cP_\restr}(\mathbf{e})$. For notational convenience, let $q\coloneqq \proj_{\cP_\restr}$.

For $R_i^{q(\mathbf{e})}\simiid M_{q(\mathbf{e})}$, our final relaxation of our optimization problem is 
\vspace{-1ex}
\begin{align}\label{eq:relaxed_obj}
\argmin_{\mathbf{e}\in \cE^d} \frac{1}{n}\sum_{i=1}^n \ell\left(R_i^{q(\mathbf{e})}\right). 
\end{align}
Given the optimum $\mathbf{e}^{\star}$, the final designed prompt is obtained by $\proj_{\cP_\restr}(\mathbf{e}^{\star})$. 
\subsection{Optimization Approaches} \label{sec: opt approaches}
We address the minimization problem in \autoref{eq:relaxed_obj} as a black-box optimization, naturally leading to the application of standard techniques to find optimal prompts. We utilize two black-box optimization techniques, the Square Attack algorithm \citep{andriushchenko2020square}, typically employed for generating black-box adversarial examples, and recent advancements in high-dimensional Bayesian optimization \citep{eriksson2021high}. 

\subsubsection{Square Attack}\label{sec: square attack}

The first black box optimization algorithm we consider is the Square Attack algorithm \citep{andriushchenko2020square}. Although originally created for attacking image classifiers, its simplicity and competitive query complexity make it a promising candidate for finding adversarial prompts. 

The Square Attack algorithm is an iterative algorithm to optimize an arbitrary high dimensional black-box $f$ with solely function evaluations. For simplicity, assume $f:\R^d\to \R$ and we would like to find an input $X$ that achieves a low value for $\ell$.

Let $x_{t}$ denote the candidate vector at iteration $t$ and $x_0$ be a chosen initialization. The iteration update for the Square Attack comprises three steps:
\begin{enumerate}
    \item \textbf{Subset Selection:} Select a subset $S\subseteq [d]$ of the indices of $x_{t}$ to update. Let $x^{(S)}_{t}$ be the corresponding subsetted vector of $x^t$. 
    
    \item \textbf{Sample Values:} Sample $v_1,\ldots,v_k \in \R^{\vert S\vert}$ and create $k$ new candidate vectors modifying only the entries in $S$ to $v_1,\ldots, v_k$: $x^{(S)}_{t,i}\coloneqq v_i$ and $x^{(\bar{S})}_{t,i}\coloneqq x^{(\bar{S})}$ for $i\in [d]$, where $\bar{S}$ is the complement of $S$.
    
    \item \textbf{Update:} Update $x_{t+1}$ with the vector attaining the lowest value among $f(x_{t,1}),\ldots, f(x_{t,k})$ and the original $f(x_{t})$.
\end{enumerate}

In our implementation, for the Subset Selection step, we choose a random subset $S$ where $\vert S\vert\approx d/10$. For the Sample Values step, we choose a random sample using a normal distribution centered around the previous value $x_{t}^{(S)}$ and identity covariance matrix weighted by a constant term $c$ and the standard deviation of the previous iteration's function evaluations. The constant term $c$ is a proxy for the `step size', we select $c$ by dividing the average distance between embedding vectors by $10$ (we choose $c=0.1$). We find that normalizing by the standard deviation improves performance, since we are less likely to take large steps close to convergence. More precisely, 
    \[v_i\simiid \frac{c}{\mathrm{stdev}(f(x^{(S)}_{t,1}),\ldots,f(x^{(S)}_{t,k}))}\cdot \cN\left(x_{t}^{(S)},\mathbf{I}\right).\]
We provide explicit algorithm details in \Cref{algo: square attack}. In our setting, we choose $\ell$ for our optimization function $f$ and $X$
is our flattened word embedding in $\R^{m\cdot d}$.

\subsubsection{Bayesian Optimization}
Sample efficiency is critical when targeting large foundation models, as evaluating the loss in \autoref{eq:relaxed_obj}, even for $n=5$, can take several minutes. Bayesian Optimization (BO) is a general purpose method for solving noisy black-box optimization problems~\cite{snoek2012practical,frazier2018tutorial}, especially when sample efficiency is desired. 

In Bayesian optimization, one is given a (possibly initially empty) set of function evaluations $\mathcal{D} : \left\{(x_{1},y_{1}),\ldots,(x_{t},y_{t})\right\}$, with $y_{i}$ a noisy observation of $f(x_{i})$. A probabilistic surrogate model--commonly a Gaussian process (GP) \cite{rasmussen2003gaussian}--is trained on this dataset to obtain a predictive model of the objective function, $p(y^{*} \mid x^{*}, \mathcal{D})$.

An \textit{acquisition function} is used which leverages this predictive posterior to find the most promising candidates in the search-space to evaluate next, efficiently trading off exploration and exploitation. When new data is acquired, the surrogate model is updated and becomes progressively more accurate. By sequentially selecting candidates to evaluate in this manner, BO can reduce the number of evaluations needed to optimize expensive black-box functions. For a more in-depth introduction to Bayesian optimization, see \citep{garnett_bayesoptbook_2023}.

\paragraph{Trust Region Bayesian Optimization (TuRBO).} Since we may have a token embedding dimension up to $m=768$ and optimize over $d=4$ tokens, we may search over a large $768 \times 4 = 3072$-dimensional space for optimal prompts, a dimensionality well out of reach for traditional Bayesian optimization methods. \citet{turbo} propose Trust Region Bayesian Optimization (TuRBO) which has enabled the use of BO on much higher dimensional functions without making additional assumptions about the search space.

TuRBO mitigates the curse of dimensionality which typically plagues BO algorithms in high-dimensional spaces by dynamically limiting the search space to be within a hyper-rectangular \textit{trust region} so that BO avoids over-exploring the exponentially large search space $\mathcal{X}$. 

The trust region is a hyper-rectangular subset of the input space $\mathcal{X}$ centered at the best point found by the optimizer--the \textit{incumbent}--$\bx^{+}_i$ and has a side-length $\beta_i \in [\beta_{min}, \beta_{max}]$. If a local optimizer improves upon its own incumbent $\rho_{succ}$ times in a row, $\beta_i$ is increased to $\min(2\beta_i, \beta_{max})$. Similarly, when a local optimizer fails to make progress $\rho_{fail}$ times in a row, the length $\beta_i$ is reduced to $\beta/2$. If $\beta_i < \beta_{min}$, that local optimizer is restarted.

%% file: loss_threats.tex
The provided formulation for adversarial prompts is flexible and can accommodate a variety of prompt classes and loss functions. The specific choice of prompt class $\cP_\restr$ determines the so-called \emph{threat model} from adversarial examples, or the space of tokens that the adversary is allowed to use to construct a prompt. On the other hand, the loss function $\ell$ specifies the targeted behavior that the adversary wishes to induce in the generated outputs. Specifying both the allowable prompt class and loss function fully defines both the threat model and the goal of the adversarial prompt. 

\subsection{Adversarial Targets}
In contrast to the standard classification setting for adversarial examples, adversaries could have an arbitrarily complex goal when searching for prompts. To actualize this unrestricted space, we consider the following two candidate loss functions that provide concrete, measurable goals for the adversary. 

\paragraph{Classifier Loss.} 
One way to provide a target for adversarial prompts is to use a pretrained classifier. An adversary can then try to find a prompt that pushes the generated text or images toward a particular class. Specifically, the classifier loss is the negative log probability of a class $\CLS$ (e.g. \texttt{cat}, \texttt{dog}, \texttt{car}, etc.) from a classifier $g$: 
\vspace{-2ex}
\begin{align}
\label{eq:vision_loss}
    \ell(R) = -\log g(R)_\CLS.
\end{align}
With this loss, the adversary aims to force the model $M_p$ to generate images of the target class $\CLS$. For example, for text-to-image models, an adversary could use an ImageNet classifier to find adversarial prompts that generate images of dogs. For text-to-text models, an adversary could use a sentiment prediction model to prompt the model into generating text with positive or negative sentiment.  

\paragraph{Feature Loss.} 
We also consider a second loss function that captures more granular information using features in the generated outputs. For example, when $R$ is generated text, $\ell(R)$ could be the \textit{perplexity} of the text, computed using a separate language model. An adversary would then use this loss to find a prompt that pushes the model into generating text with high perplexity, perhaps nonsense or gibberish text.

\subsection{Threat Models} \label{sec: threat models}
Similar to adversarial examples for discriminative models, an adversarial prompt needs to have a well-defined threat model that cleanly states what an adversary is and is not allowed to do. However, existing threat models in language typically require a clean example that serves as a center of allowable perturbations for the adversary (i.e. by perturbing characters and words or paraphrasing). These threat models do not directly carry over to the prompting setting, as there is no clean example to center around.

Thus in this paper, we consider the following three types of threat models to restrict the adversary. 
\begin{enumerate}
    \item \label{task: unrestricted prompts 1} 
    \textbf{Unrestricted Prompts}: The adversary is completely unrestricted $(\cP_\restr = \cP)$ and can use any sequence of tokens. 
    \item \label{task: restricted prompts 2} \textbf{Restricted Prompts}: The adversary is restricted to prompts that only use a subset of tokens $\cT_\restr \subset \cT$. This restricted prompt space ($\cT_\restr^d = \cP_\restr \subset \cP$) eliminates prompts that trivially achieve low loss, such as the prompt containing only the class name. 

    Specifically in the Classifier Loss setting, $\cT_\restr$ excludes tokens that are similar to the desired class $\CLS$. For example, if we would like to generate images of dogs, then $\cT_\restr$ would exclude tokens such as \texttt{puppy}, \texttt{labrador}, and emojis of dogs.
    
    \item \label{task: restricted prepending prompts 3} \textbf{Restricted Prepending Prompts}: The adversary is restricted to a subset of tokens and must prepend them to a predefined prompt $p'$. More specifically, $\cP_{\textrm{pre}}=\{\texttt{concat}(p_1,p'):p_1\in\cP_{\restr}\}$ where $\cP_\restr$ is the set of restricted prompts in Task \ref{task: restricted prompts 2}. 
    
    For example, the predefined prompt could be $p'$ = \texttt{a picture of a \CLS{}}. The adversary prepends tokens to cause the model to generate images of a different class $\texttt{CLS}_2$ (as specified in the loss function), while excluding tokens related to $\texttt{CLS}_2$ as in the previous task.
\end{enumerate}

These three threat models increase in constraint and difficulty for the adversary. In particular, the third task has opposing text in the prompt by construction that an adversarial prompt must overcome while not using related words.

%% file: experiments.tex
Utilizing the Token Space Projection discussed in \Cref{sec: token space projection}, we apply the two black-box optimization methods discussed in \Cref{sec: opt approaches} to each task introduced in \Cref{sec: threat models}. For the text-to-image tasks, we use embeddings from CLIP ViT-L/14 tokenizer \citep{Radford2021LearningTV} where $m=768$, and optimize over $d=4$ tokens. For the text-to-text tasks, we use embeddings from the L-2 H-128 BERT model \citep{Turc2019WellReadSL} where $m=128$, and optimize  over $d=6$ tokens.
Code to reproduce all results is available at \url{https://github.com/DebugML/adversarial_prompting}. Implementation details and hyperparameters are detailed in Appendix~\ref{app:implementation}.

\subsection{Image Generation}
We consider the optimization Tasks \ref{task: unrestricted prompts 1}, \ref{task: restricted prompts 2}, and \ref{task: restricted prepending prompts 3} as described in \Cref{sec: loss and threats} using the Classifier Loss. We use the Stable Diffusion v1.5 model \citep{StableDiff} to generate 512x512 images, using a classifier free guidance scale of 7.5 \citep{ho2021classifierfree} and 25 inference steps with the default scheduler. For our classifier loss function, we use the standard ResNet18 from TorchVision \citep{torchvision2016} and use the negative log class probability for the desired class.

For Tasks \ref{task: unrestricted prompts 1} and \ref{task: restricted prompts 2}, we allow the black-box optimization methods to query a total of 5,000 prompts before terminating. If the optimizer fails to make progress for 1,000 prompt consecutive queries, we terminate early. Since Task \ref{task: restricted prepending prompts 3} is more difficult, we allow 10,000 prompts to be queried and terminate early after failing to make progress for 3,000 consecutive queries.

\subsubsection{Quantitative Evaluation Metrics}
\label{sec:success-methods}
\input{results_table.tex}
In addition to example images demonstrating success, we measure the \textit{success} of optimized prompts quantitatively in two ways.

\paragraph{Most Probable Class Success (MPC Success).} The optimizer achieves \textit{MPC success} if it finds a prompt for which the target class is the largest probability class for the majority of 5 images generated according to a pretrained ResNet18 classification model.

\paragraph{Out-Performing Simple Baseline Prompts (OPB Success).} Denote by \CLS{} the target ImageNet class. We compute the loss using the following two baseline prompts: \CLS{} and \texttt{a picture of a \CLS{}}. The optimizer achieves \textit{OPB success} if the optimizer finds a prompt that achieves lower loss than both of these baseline prompts.

\begin{figure*}[t]
\centering
\begin{subfigure}[b]{0.48\textwidth}
\addtolength{\tabcolsep}{-5.5pt}   
\begin{tabular}{lll}
\multicolumn{3}{l} {\fontsize{11pt}{13pt}\textbf{Goal Class:} \texttt{lizard}}\\
     \multicolumn{3}{l} {\ttbfs{louisiana argonhilton deta}}\\
     \includegraphics[width=0.333\columnwidth]{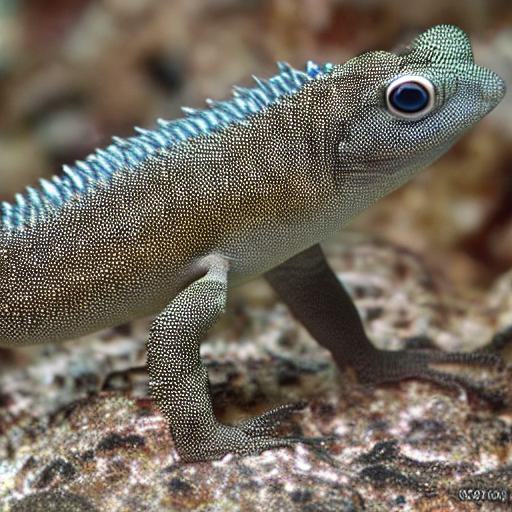} &
     \includegraphics[width=0.333\columnwidth]{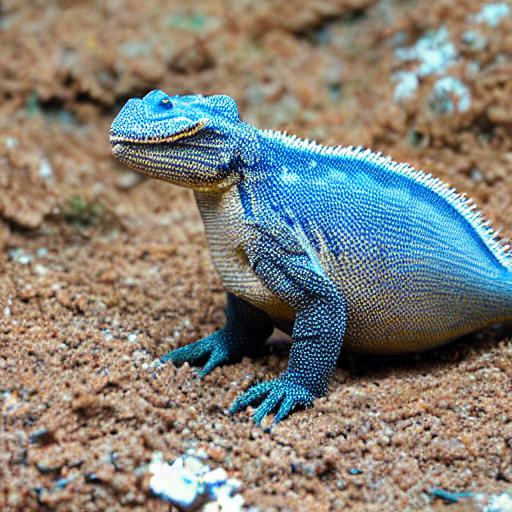} &
     \includegraphics[width=0.333\columnwidth]{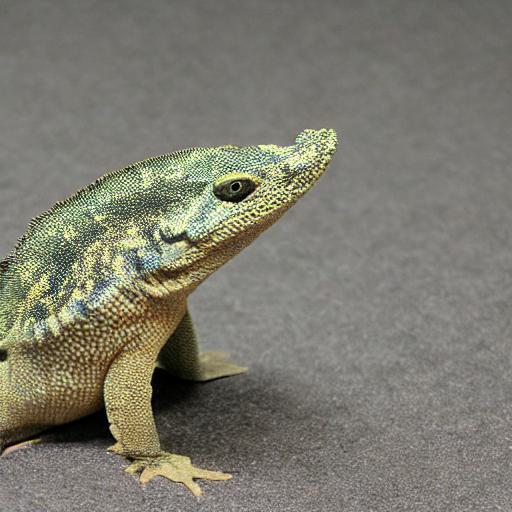}
\end{tabular}
\newline
\begin{tabular}{llll}
    \ttbfs{louisiana} & \ttbfs{argon} & \ttbfs{hilton} & \ttbfs{deta} \\
    \includegraphics[width=0.25\columnwidth]{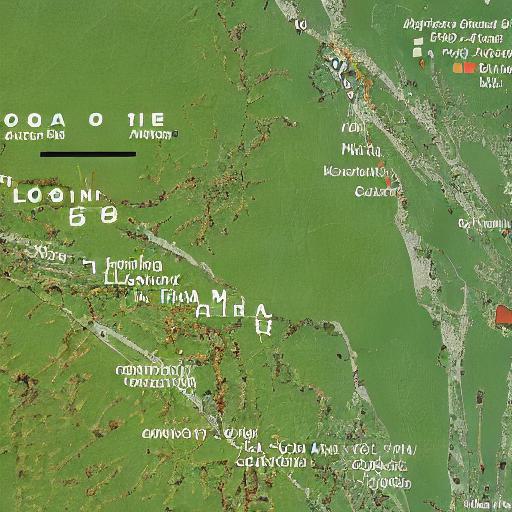} &
    \includegraphics[width=0.25\columnwidth]{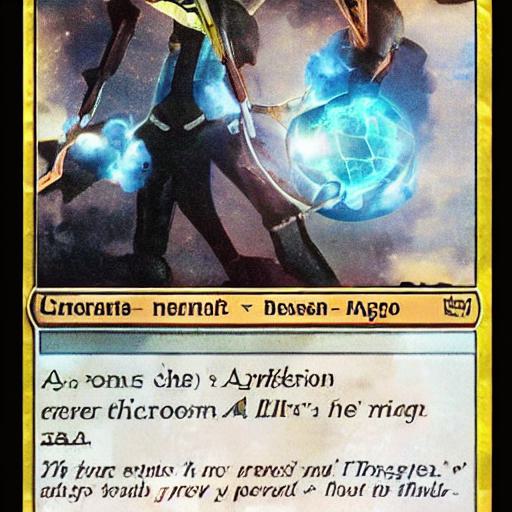} &
    \includegraphics[width=0.25\columnwidth]{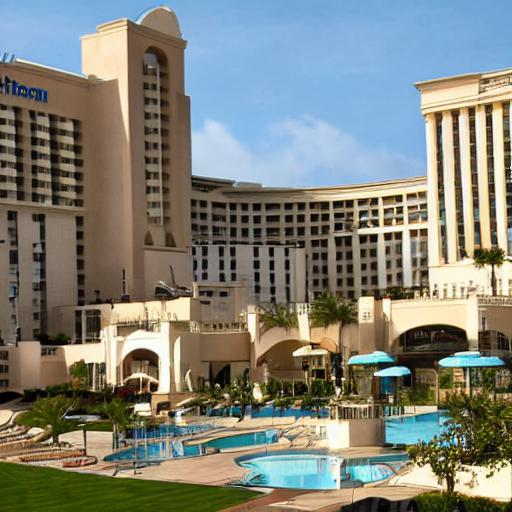} & 
    \includegraphics[width=0.25\columnwidth]{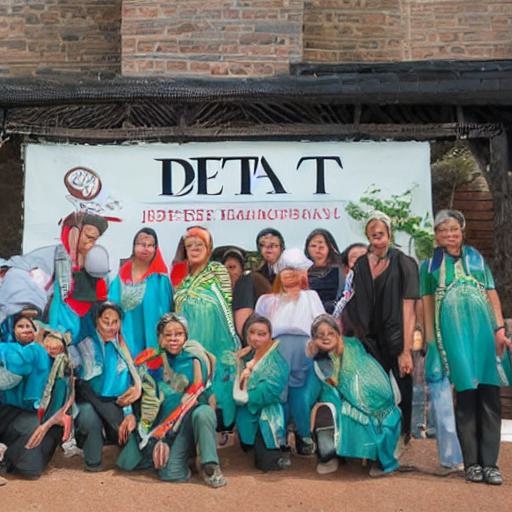}
\end{tabular}

\caption{Task \ref{task: restricted prompts 2} generated images with the class \texttt{lizard}.}
\label{lizard-and-lipstick}
\end{subfigure}
\hfill
\begin{subfigure}[b]{0.48\textwidth}
\centering
\addtolength{\tabcolsep}{-5.5pt}   
\begin{tabular}{lll}
    \multicolumn{3}{l} {\fontsize{11pt}{13pt}\textbf{Goal Class:} \texttt{aircraft}}\\
     \multicolumn{3}{l} {{\fontsize{8pt}{10pt}\ttbfs{pegasus yorkshire wwii taken a picture of the ocean}} {} }\\
     \includegraphics[width=0.333\columnwidth]{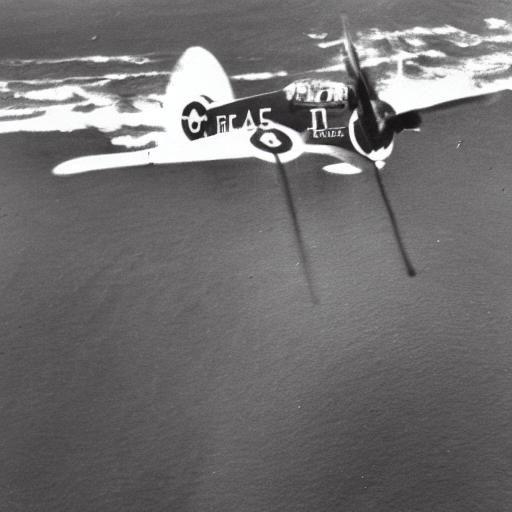} &
     \includegraphics[width=0.333\columnwidth]{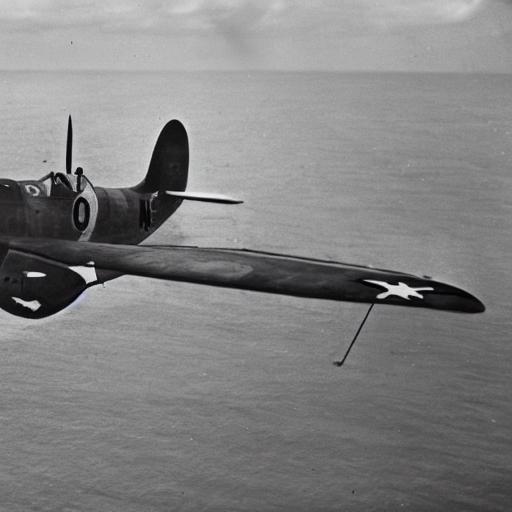} &
     \includegraphics[width=0.333\columnwidth]{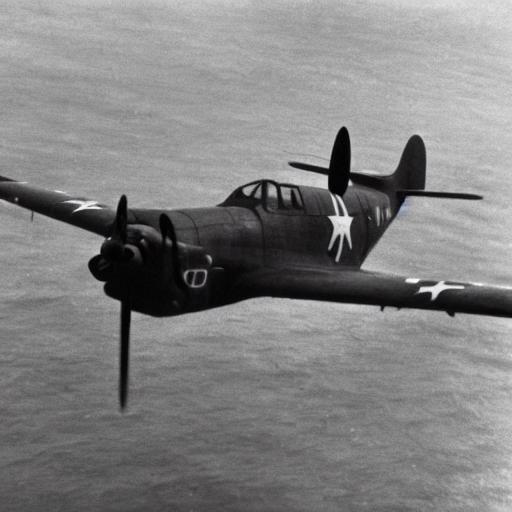}
\end{tabular}
\newline
\begin{tabular}{llll}
    \ttbfs{pegasus} & \ttbfs{yorkshire} & \ttbfs{wwii} & \ttbfs{taken} \\
    \includegraphics[width=0.25\columnwidth]{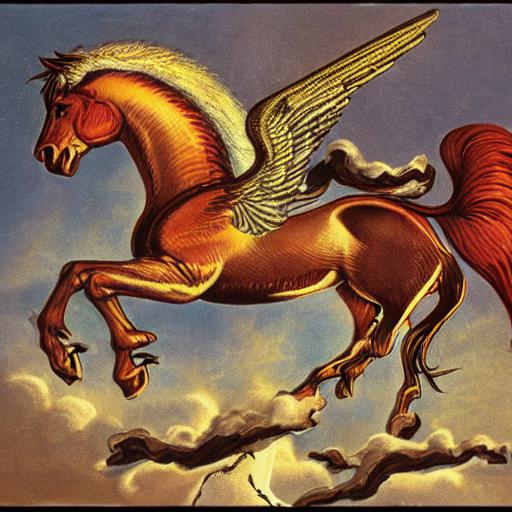} &
    \includegraphics[width=0.25\columnwidth]{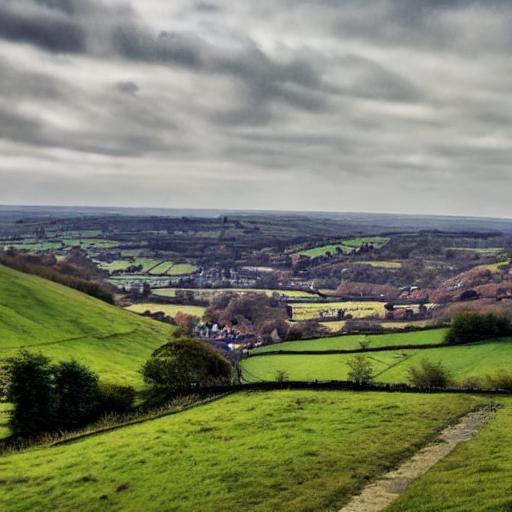} &
    \includegraphics[width=0.25\columnwidth]{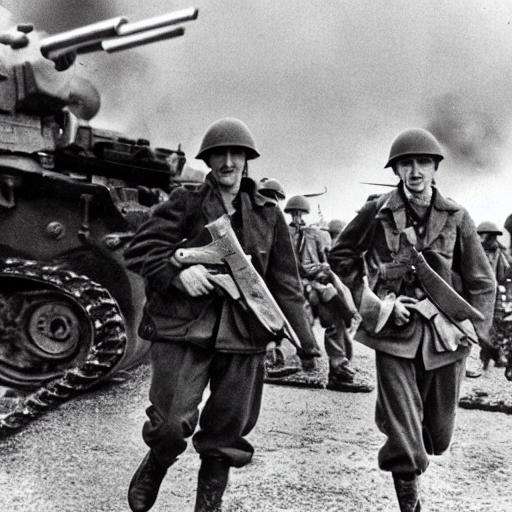} & 
    \includegraphics[width=0.25\columnwidth]{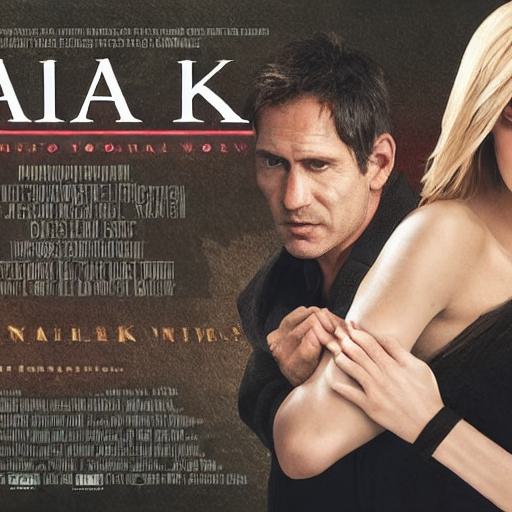}
\end{tabular}
\addtolength{\tabcolsep}{5.5pt} 
\caption{Task \ref{task: restricted prepending prompts 3} generated images with the class \texttt{aircraft}.}
\label{aircraft-and-ballplayer-prepend}
\end{subfigure}
\caption{Sample images generated using adversarial prompts. In both figures, the first row contains images generated by the optimized prompt and are classified by the ResNet18 classifier as the target class. 
The second row contains images generated from each individual token in the optimized prompt, none of which are classified as the target class. See Appendix \ref{app: additional examples} for more examples.}
\vspace{-3ex}
\end{figure*}
\subsubsection{Unrestricted Prompts}
\label{sec:manyimnet}

We consider a set of 400 target classes randomly selected from ImageNet classes that have a single token class name (e.g., excluding classes like \texttt{brown bear}). We report success rates for these 400 classes in \Cref{tab: results}. Both methods find prompts that are successful under both definitions from many target classes, with TuRBO consistently achieving a high rate of success. 

\paragraph{Higher-Level Classes.} In addition to the 400 regular ImageNet target classes, we randomly select a set of 150 higher-level classes from the ImageNet hierarchy. Higher-level ImageNet classes encompass some larger number of standard classes (i.e. \texttt{wolf} is a higher-level class encompassing the base-classes: \texttt{timber wolf}, \texttt{white wolf}, \texttt{red wolf}, \texttt{coyote}). For each higher-level class, the optimizer seeks a prompt that generates images that fall in any of the associated base-classes. The probability that an image falls in a higher-level class is the maximum over the probabilities of all associated base-classes. 

Success rates for the 150 higher-level classes are reported in \Cref{tab: results}. Optimizing for higher-level classes is more straightforward since the optimizer does not need to differentiate between highly related classes such as \texttt{timber wolf} and \texttt{white wolf}. We observe that success rates are therefore uniformly higher in this setting.

\subsubsection{Restricted Prompts}
\label{sec:restricted}

In \Cref{sec:manyimnet}, the optimization methods used an unconstrained vocabulary when designing prompts. Thus, when generating pictures of a wolf, the optimizers were allowed to use the token \texttt{wolf}, as well as other highly related words. In this section, we make this task significantly more challenging by forbidding the optimizers from using certain \textit{high-similarity tokens} in the output prompt.

We randomly select 12 target regular ImageNet classes and another 20 target higher-level ImageNet classes.
Since the restricted prompt optimization task is more difficult, we select these target classes from the subset of classes for which MPC success was achieved on the unrestricted task. 
The 12 selected regular ImageNet classes are listed in \Cref{tab:lower-classes-list}. The 20 selected higher-level ImageNet classes can be found in \Cref{tab:higher-classes-list}. 

We define a high-similarity token (HST) as a token $t$ where the prompt $p= t$ achieves low loss. We exclude all tokens which achieve a log probability of the target class higher than $-3.0$. In order to determine a comprehensive list of HSTs for each target class, we compute the loss for each of the 50k tokens available to the model. Precomputing the loss for these 50k tokens is computationally prohibitive-- thus we restrict our experiments to 32 target classes. We report results for optimization without access to HSTs for the 32 selected ImageNet classes in \Cref{tab: results}. Example optimal prompts with generated images can be found in \Cref{lizard-and-lipstick}, with additional results in \Cref{app: additional examples}. 

\subsubsection{Restricted Prepending Prompts}

We consider the prepending task introduced in Task \ref{task: restricted prepending prompts 3} of \Cref{sec: threat models} on a subset of the 32 classes used in \Cref{sec:restricted}. We prepend prompts onto the following three strings: \texttt{a picture of a dog}, \texttt{a picture of a mountain}, and \texttt{a picture of the ocean}. We provide quantitative results for all three variations of this task in \Cref{tab: results}. Because the prompts \CLS{} and \texttt{a picture of a \CLS{}} are no longer necessarily strong baselines when prepended to e.g. \texttt{a picture of a dog}, we no longer report OPB success for this task. Example adversarial prompts with generated images can be found in \Cref{aircraft-and-ballplayer-prepend}, with additional results in Appendix \ref{app: additional examples}. 

\subsection{Text Generation} 
\begin{figure}[t]
    \centering
    \includegraphics[width=0.9\columnwidth]{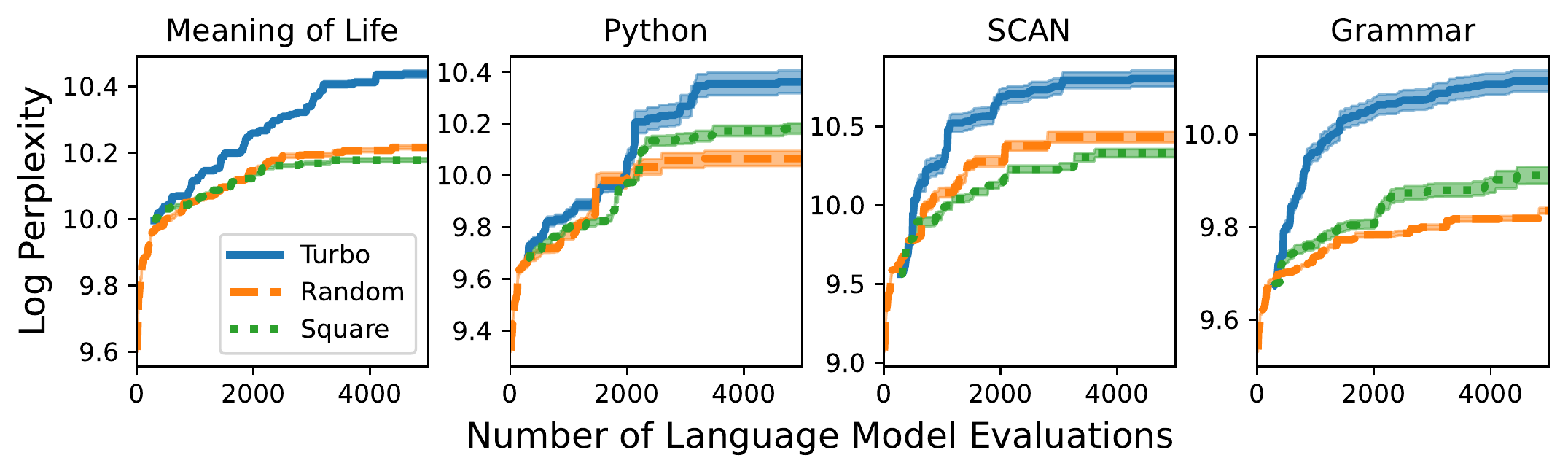}
    \caption{Plots of the maximum log perplexity with respect to the number of Vicuna 13B-v1.1 evaluations. The values are averaged over 20 random seeds, with the one standard error interval shaded. The full seed prompts are in \cref{tab: seed prompts}.}
    \label{fig: text perplexity plot}
    \vspace{-2ex}
\end{figure}

With the rise in the use of generative text models, it's imperative to embed safety measures, ensuring their outputs align with human intentions while avoiding potential risks. We consider the relatively innocuous task of optimizing prompts for large language models to increase the perplexity of the output. Perplexity quantifies the overall level of likelihood of a sequence, measured by the negative log-likelihood as measured by a language model -- we use GPT-2 \citep{gpt2}. For a particular prompt, our objective is the average perplexity of the middle three of five generated outputs.

We use the Stable Vicuna 13B-v1.1 model \citep{vicuna2023}--a state of the art freely available LLM trained with human feedback--for text generation with a maximum generated length of $50$ tokens. We optimize perplexity with a maximum budget of 5,000 prompts, terminating early if no progress is made for 2,000 prompts in a row.

We optimize over prompts to prepend to a seed prompt (which we denote as `Meaning of Life', `Python', `SCAN', `Grammar'), and replicate these experiments 20 times per seed prompt. To choose seed prompts, we randomly selected prompts from the GPT4ALL and PromptSource datasets~\citep{gpt4all,bach2022promptsource} -- see \cref{tab: seed prompts} in \cref{app: text gen} for a list of all prompts. We use the Square Attack and TuRBO optimizers from \cref{sec: opt approaches} and also include a random search baseline, where we simply randomly sample tokens. We provide the performance of these approaches on four seed prompts in \cref{fig: text perplexity plot}, and more results in \cref{app: text gen}.

\begin{figure}
\centering
 \tbox[\textbf{Prompt: }\textit{intent oder ranch delegate source every} Explain list comprehension in Python.]{
            \textbf{Generated Examples: } \\
            \begin{tabular}{l p{12cm}}
                 1. &  Willkommen auf meinem YouTube-Kanal! Hier ist mein erstes Video, in dem ich dir erkläre, was Intent oder Ranch ist und wie man es in Python verwenden kann.\\
                 2. & Willkommen auf meinem GPT-3-Konto! Ich bin ein künstlicher Intelligenz-Sprachmodell und kann dir bei deinen Frag en helfen.\\ 
                 3. & Willkommen auf meinem ZDF-Zukunftsforum-Kanal! Hier werden wir uns mit den wichtigsten Themen der Zukunft ausein andersetzen und diskutieren, welche Entwicklungen unser Leben
            \end{tabular}
        }
    \caption{Three examples of text generated from Vicuna 13B-v1.1 using our optimized prompt in italics and the original prompt to which we prepend. The generated text are nonsensical or irrelevant to the prompt.}
\label{fig: text-examples}
\vspace{-3ex}
\end{figure}
\cref{fig: text perplexity plot} demonstrates our ability to  increase the perplexity of the generated text, often by an order of magnitude. We also provide an example of the optimized adversarial prompt for the `Python' seed prompt in \cref{fig: text-examples}. This serves as a preliminary evidence that we may successfully optimize adversarial prompts to a desired criterion.

%% file: results_table.tex
\begin{table}[]
\centering
\caption{Success rates for TuRBO and Square Attack for all image generation tasks. We report the proportion of these classes for which each method achieved success according to our two measures of success discussed in \Cref{sec:success-methods}. Both refers to MPC and OPB. Single refers to single/regular image-net classes. Higher refers to image-net classes further up in the image-net class hierarchy.}
\begin{tabular}{clcc cc}
\toprule
Task                                                                             & \multicolumn{1}{l }{Metric} & \begin{tabular}[c]{@{}c@{}}TuRBO\\ (Single)\end{tabular} & \begin{tabular}[c]{@{}c@{}}Square\\ (Single)\end{tabular} & \begin{tabular}[c]{@{}c@{}}TuRBO\\ (Higher)\end{tabular} & \begin{tabular}[c]{@{}c@{}}Square\\ (Higher)\end{tabular} \\ 
\midrule
\multirow{3}{*}{\begin{tabular}[c]{@{}c@{}}Task 1\end{tabular}} & MPC                        & 18.5\%                                                   & 1.5\%                                                     & 42.0\%                                                   & 12.7\%                                                    \\
                                                                                 & OPB                        & 36.8\%                                                   & 13.5\%                                                    & 61.3\%                                                   & 23.3\%                                                    \\
                                                                                 & Both                    & 10.5\%                                                   & 0.0\%                                                     & 36.0\%                                                   & 7.3\%                                                     \\ \midrule
\multirow{3}{*}{\begin{tabular}[c]{@{}c@{}}Task 2\end{tabular}}   & MPC                        & 58.3\%                                                   & 8.3\%                                                     & 60.0\%                                                   & 0.0\%                                                     \\
                                                                                 & OPB                        & 83.3\%                                                   & 58.3\%                                                    & 75.0\%                                                   & 70.0\%                                                    \\
                                                                                 & Both                   & 50.0\%                                                   & 8.3\%                                                     & 55.0\%                                                   & 10.0\%                                                    \\ \midrule
\begin{tabular}[c]{@{}c@{}}Task 3\end{tabular}           & MPC                        & 25.0\%                                                   & 0.0\%                                                     & 37.3\%                                                   & 0.0\%           \\\bottomrule  \\
\end{tabular}

\label{tab: results}
\vspace{-3ex}
\end{table}

%% file: related_work.tex
\paragraph{Adversarial Attacks.} 
The field of adversarial examples has a long history of developing methodologies for finding input perturbations that can change the output of a classifier \citep{biggio2018wild}, starting from early text-based attacks to circumvent spam filters \citep{dalvi2004adversarial,lowd2005adversarial, lowd2005good}. Gradient-ascent based approaches have long been used to generate small perturbations with bounded $\ell_p$ noise \citep{biggio2013evasion, szegedy2013intriguing, goodfellow2014explaining, tramer2019adversarial,maini2020adversarial}, as well as perceptually similar attacks \citep{laidlaw2020perceptual}. In computer vision, visually realistic adversarial attacks have been produced on glasses \citep{sharif2016accessorize}, clothing \cite{wu2020making}, printable patches \citep{eykholt2018physical, chen2018shapeshifter, liu2018dpatch, thys2019fooling}, and more \citep{engstrom2017rotation,wong2019wasserstein}. In language, word embedding \citep{miyato2016adversarial}, word substitution \citep{jia2019certified}, semantically similar \citep{alzantot2018generating}, and BERT-based \cite{li2020bert} attacks were developed to flip the output of a classifier. Most recently, the researchers have begun designing prompt attacks on LLMs such as prompt injection \citep{perez2022ignore} and prompt backdoors \citep{xu2022exploring} that work in specific scenarios. Concurrently, \citet{yang2023sneakyprompt} developed a reinfocement learning-based framework for circumventing safety filters on text-to-image generative models. 

Of particular relevance are black box adversarial attack frameworks, those which require only query-access to the model being attacked. Early attacks required training a substitute model to generate adversarial examples \citep{papernot2017practical}. Other works have developed zeroth order optimization techniques to avoid training a substitute model \citep{narodytska2016simple, chen2017zoo}, with later improvements to reliability and query complexity \citep{brendel2017decision, ilyas2018black, cheng2018query, andriushchenko2020square}. Other variants employ tools ranging from combinatorial optimization \citep{moon2019parsimonious}, bandits \citep{ilyas2018prior}, and confidence scores \citep{guo2019simple}. Most of these frameworks have been demonstrated primarily in finding $\ell_p$ adversarial examples in the vision setting.  
\vspace{-1ex}
\paragraph{Prompting.} 
Recent years have seen a prevalence of prompting \citep{liu2023pre} as a way to interact with models instead of fine-tuning with data. With a large enough of a language model, it became possible to adapt LLMs to new tasks with just a natural language instruction and a few examples as input \citep{brown2020language}. Since then, a number of specialized LLMs have been developed for various tasks including code generation \citep{chen2021evaluating} and storytelling \citep{yuan2022wordcraft}, Prompting has also seen recent use as an interface for guiding zero-shot image generation \citep{ramesh2021zero, rombach2022high} and image editing \citep{kawar2022imagic, couairon2022diffedit}. By conditioning image generation on provided text, users can generate high quality images with variable content/style.

%% file: conclusion.tex
\textbf{Broader Impacts.} This paper presents a systematic and automated mechanism for manipulating generative models into producing unintended results. We emphasize that any such method, ours or similar, is inherently a double-edged sword. It can be used to bypass alignment safeguards, effectively `jailbreaking' these models. Yet, these methods may facilitate a more systematic alignment process, an alternative to current hand-crafted efforts. Given the relative ease of manually crafting `jailbreak' prompts, we posit that the advantages of an automated approach for investigating unintended behavior in generative models currently outweigh the potential drawbacks.

As closed-source foundation models grow in popularity, it is critical to understand how they react to unexpected prompts. In this paper, we demonstrate that it is possible to efficiently and consistently manipulate the generated output of these foundation models with adversarial prompts. The attack is effective even when using only a small, four token augmentation to otherwise benign prompts. In the future, these techniques may evolve to allow practitioners greater flexibility in prompting via backend prompt engineering. For example, it may be possible to encourage or discourage certain types of outputs by augmenting a user specified prompt with a small handful of tokens found via our adversarial prompting framework.

%% file: appendix.tex
\noindent\rule{\textwidth}{1pt}
\begin{center}
\vspace{7pt}
{\Large  Appendix}
\end{center}
\noindent\rule{\textwidth}{1pt}
\section{Limitations}
The main limitations on our work are the evaluation of the open source models and necessity of real valued loss functions. Our approaches apply to arbitrary vision and language models, and we formulate our attacks as black box optimizations. However, due to the cost of systematic evaluation on black-box models, we perform our experiments on open source models, as systematically evaluating our approach on DALLE-2 or ChatGPT may become prohibitively expensive. Furthermore, our optimization requires some form of real-valued signal to ensure improvement. For possible tasks of interest, there may be a binary loss, which would possibly provide too sparse of a signal to optimize.
\section{Compute}
These experiments were evaluated on 4 Nvidia Tesla A100s. Each individual run takes about 30-60 minutes on a single GPU, for a total of about 1000 GPU hours for all experiments.
\section{Adversarial Prompt Optimization Pipeline}
\begin{figure}[ht]
    \centering
    \includegraphics[width=0.7\columnwidth]{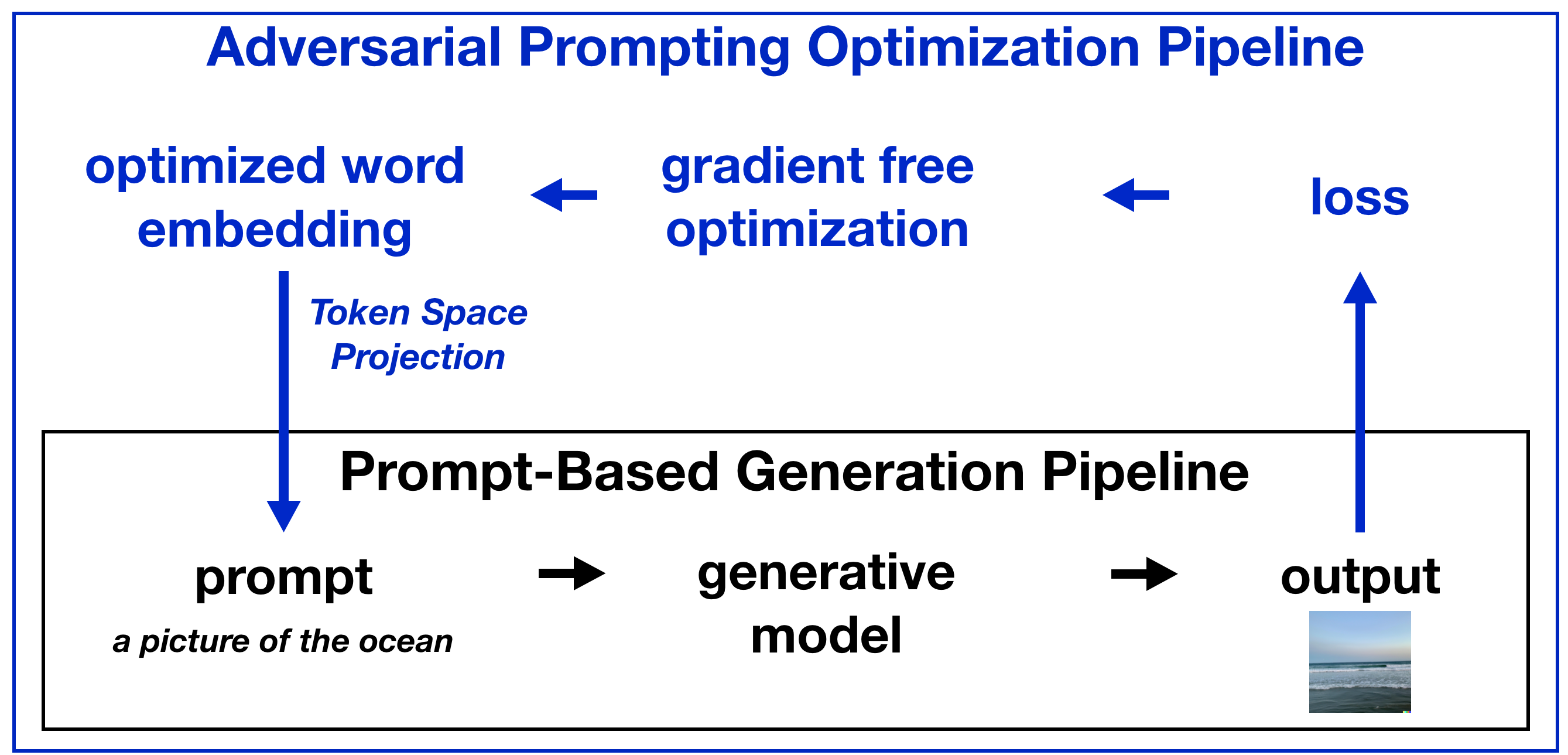}
    \caption{Adversarial prompting pipeline for a prompt-based generative model.}
    \label{fig: adv prompt pipeline}
\end{figure}
\section{Square Attack Algorithm}
We include an algorithmic description of the Square Attack described in \Cref{sec: prompt opt}.
\begin{algorithm}[h]
    \caption{Square Attack Algorithm}
    \label{algo: square attack}
    \textbf{Input: loss function $\ell:\R^d\to\R$} 
    \begin{algorithmic}[1]
        \STATE Initialize $x_0$
        \STATE $\sigma\leftarrow 1$
        \FOR{$t=0,\ldots,T-1$}
        \STATE Select random subset $S\subseteq [d]$ with $\vert S\vert = d/10$
        \STATE $x_{t,1},\ldots,x_{t,k}\leftarrow x_{t}$
        \STATE Sample $v_1,\ldots,v_k\simiid \frac{1}{10\sigma}\cN(x^{(S)}_0,I)$
        \STATE $x^{(S)}_{t,i} \leftarrow v_i$
        \STATE Compute and store $q_t \leftarrow (f(x_{t,1}),\ldots,f(x_{t,k}))$
        \STATE $\sigma \leftarrow \mathrm{stdev}(q_t)$
        \IF{$f(x_{t})\ge \min(q_t)$}
            \STATE $x_{t+1}\leftarrow \argmin q_t$
            \ELSE
            \STATE $x_{t+1}\leftarrow x_t$
            \ENDIF
        \ENDFOR
        \STATE Return $x_T$
    \end{algorithmic}
\end{algorithm}

\subsection{Selected ImageNet Classes for Tasks 2 and 3 }
\autoref{tab:lower-classes-list} lists the 12 ImageNet selected for optimizing without access to HSTs and for the prepending task. \autoref{tab:higher-classes-list} lists the 20 classes selected that were higher level in the ImageNet hierarchy.

\begin{table}[h]
\caption{The 12 regular ImageNet classes randomly selected for more difficult optimization tasks (optimizing without access to high-similarity tokens and pre-pending tasks). }
\label{tab:lower-classes-list}
\begin{center}
\begin{sc}
\begin{tabular}{p{8 cm}}
\toprule
{\large Classes} \\
\midrule
{\large tabby, sportscar, violin, ballplayer, library, lipstick, pinwheel, mask, church, goblet, goldfinch, pickup}\\
\bottomrule
\end{tabular}
\end{sc}
\end{center}
\end{table}

\begin{table}[h]
\caption{The 20 higher-level ImageNet classes selected for more difficult optimization tasks (optimizing without access to high-similarity tokens and prepending tasks).  }
\label{tab:higher-classes-list}

\begin{center}
\begin{sc}
\begin{tabular}{p{8cm}}
\toprule
{\large Classes } \\
\midrule
{\large aircraft, big-cat, bird, dog, person, truck, store, lizard, headdress, toiletry, seat, musical-instrument, piano, gymnastic-apparatus, ball, equine, swimsuit, fruit, domestic-cat, bus }\\
\bottomrule
\end{tabular}
\end{sc}
\end{center}
\end{table}
\clearpage

\section{Implementation Details and Hyperparameters}
\label{app:implementation}
We use the Hugging Face API to load in and apply our generative models, the details of which we describe in the relevant sections below. We use the open source BoTorch~\cite{balandat2020botorch,gardner2018gpytorch} implementation of TuRBO. 
Code to reproduce all results is available at \url{https://github.com/DebugML/adversarial_prompting}. For the TuRBO optimization method, all trust region hyper-parameters are set to the TuRBO defaults as used in \cite{turbo}.

Since we consider large numbers of function evaluations for several tasks, we use an approximate Gaussian process (GP) surrogate model.
In particular, we use a Parametric Gaussian Process Regressor (PPGPR) \cite{PPGPR} with a deep kernel (a GP with several linear layers between the search space and the GP kernel). We use a deep kernel with 3 linear layers with 256, 128, and 64 nodes. We use 100 random points to initialize optimization across all methods. To account for the variability in the generative model outputs, we generate 5 unique outputs for each prompt queried and compute the average loss.

\section{Additional Examples}
\label{app: additional examples}
\begin{figure}[ht]
\addtolength{\tabcolsep}{-5pt}  
\begin{tabular}{lll}
    \multicolumn{3}{l} {\fontsize{11pt}{13pt}\textbf{Goal Class:} \texttt{lipstick}}\\
     \multicolumn{3}{l} {\ttbfs{eyebrows octavia assumptions} \includegraphics[width=0.04\columnwidth]{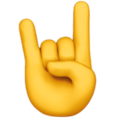}}\\
     \includegraphics[width=0.333\columnwidth]{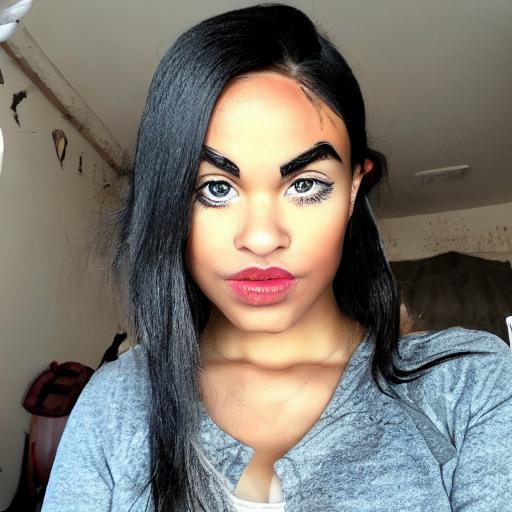} &
     \includegraphics[width=0.333\columnwidth]{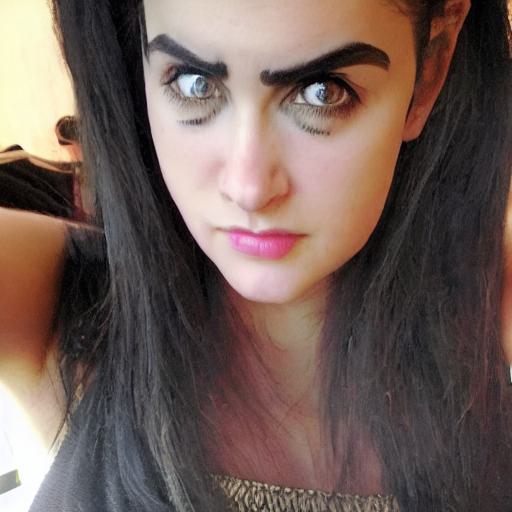} &
     \includegraphics[width=0.333\columnwidth]{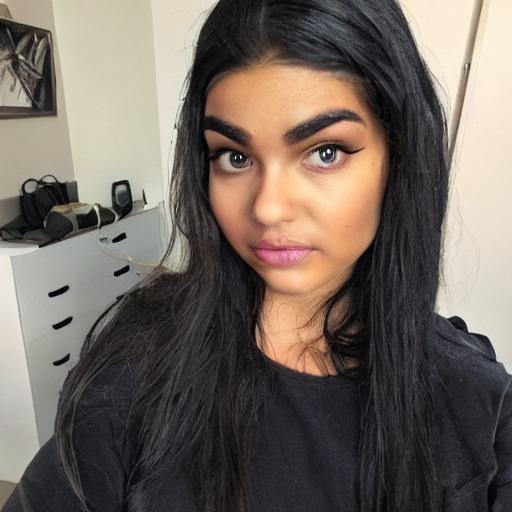}
\end{tabular}
\newline
\begin{tabular}{llll}
    \ttbfs{eyebrows} & \ttbfs{octavia} & \ttbfs{assumptions} & \includegraphics[width=0.04\columnwidth]{final_images/lipstick/sign-of-the-horns_1f918.png} \\
    \includegraphics[width=0.25\columnwidth]{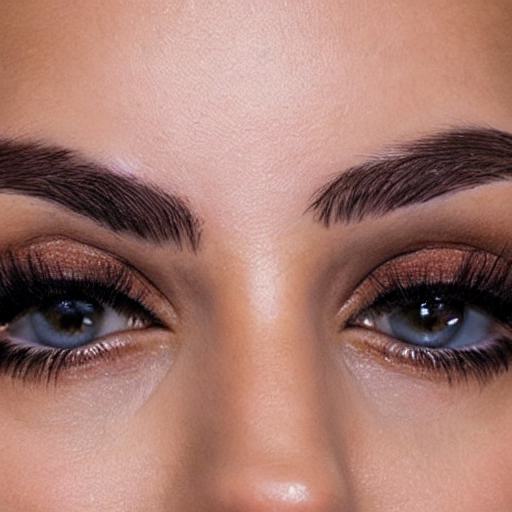} &
    \includegraphics[width=0.25\columnwidth]{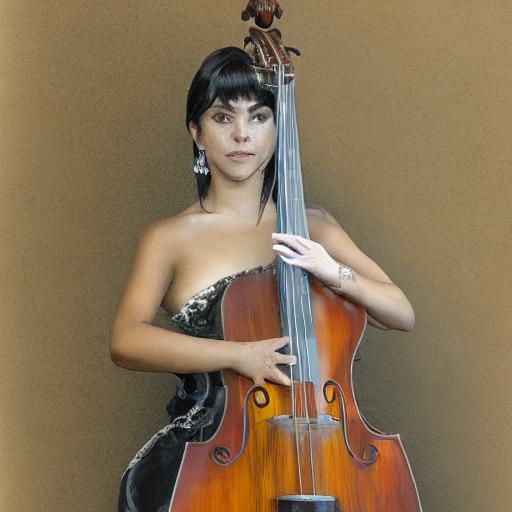} &
    \includegraphics[width=0.25\columnwidth]{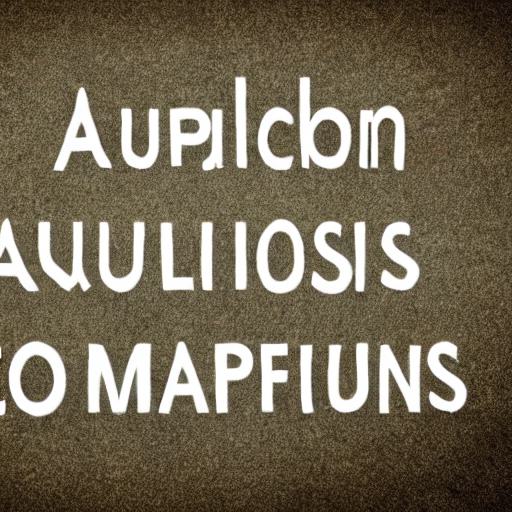} & 
    \includegraphics[width=0.25\columnwidth]{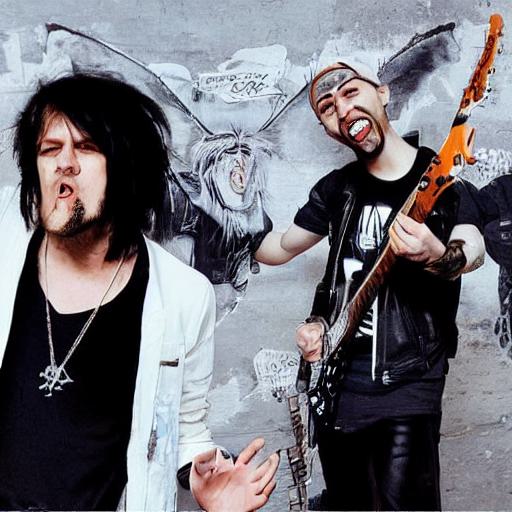}
\end{tabular}
\addtolength{\tabcolsep}{5pt}
\caption{Visualizing images generated by an optimal prompt found by TuRBO for the target higher-level ImageNet class: \texttt{lipstick}. 
Top row shows three images generated by the optimal prompt. 
All three images are classified by the ResNet as lipstick, thus indicating this was a successful prompt. 
The bottom row shows images generated from each individual token in the optimal prompt. 
None of the individual tokens are able to generate the target class.}
\label{lipstick}
\end{figure}

\begin{figure}[ht]
\addtolength{\tabcolsep}{-5pt}   
\begin{tabular}{lll}
    \multicolumn{3}{l} {\fontsize{11pt}{13pt}\textbf{Goal Class:} \texttt{ballplayer}}\\
    &&\\
     \multicolumn{3}{l} {\ttbf{mohammed} \includegraphics[width=0.02\linewidth]{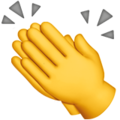} \ttbf{sal threw}}\\
     \includegraphics[width=0.333\linewidth]{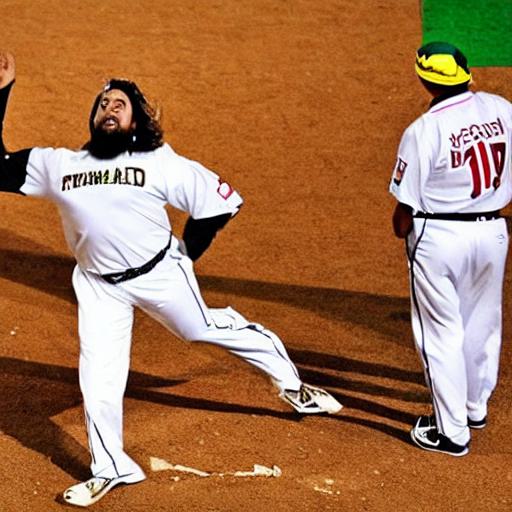} &
     \includegraphics[width=0.333\linewidth]{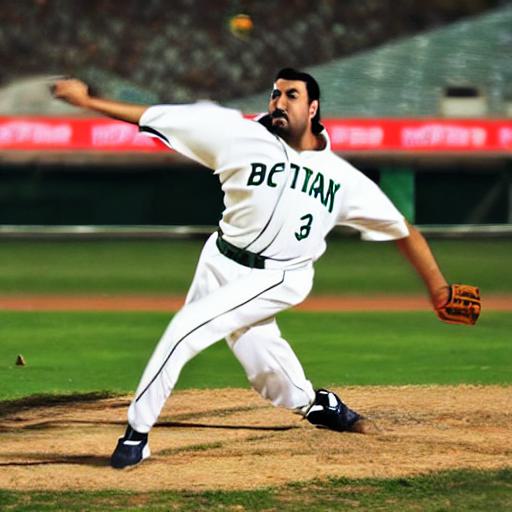} &
     \includegraphics[width=0.333\linewidth]{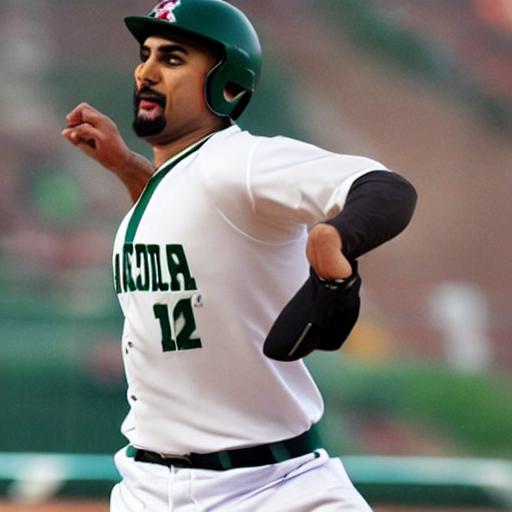}
\end{tabular}
\begin{tabular}{llll}
    \ttbf{mohammed} & \includegraphics[width=0.02\linewidth]{final_images/ballplayer/clapping-hands_1f44f.png} & \ttbf{sal} & \ttbf{threw}\\
    \includegraphics[width=0.25\linewidth]{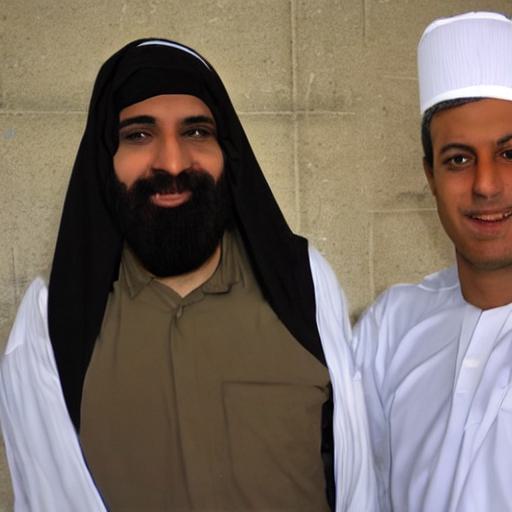} &
    \includegraphics[width=0.25\linewidth]{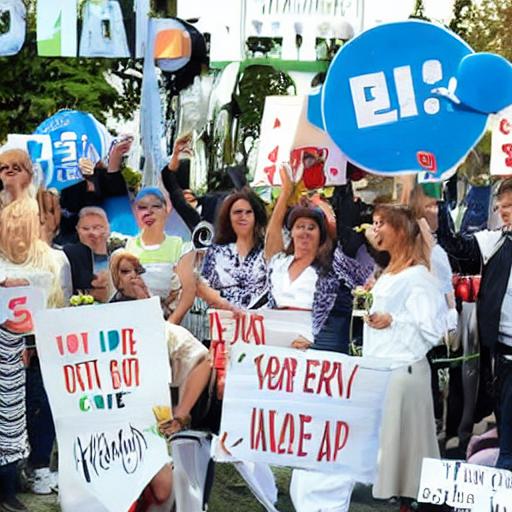} &
    \includegraphics[width=0.25\linewidth]{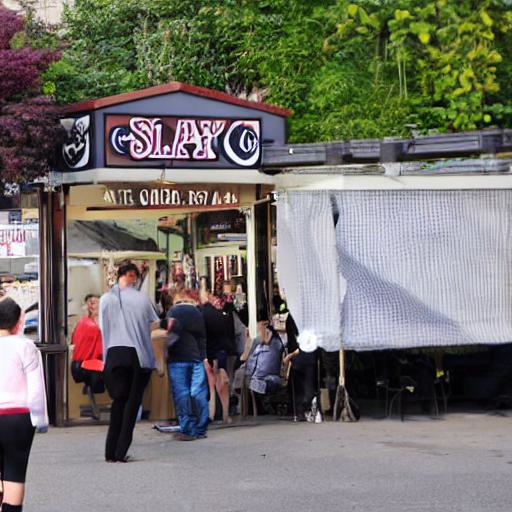} & 
    \includegraphics[width=0.25\linewidth]{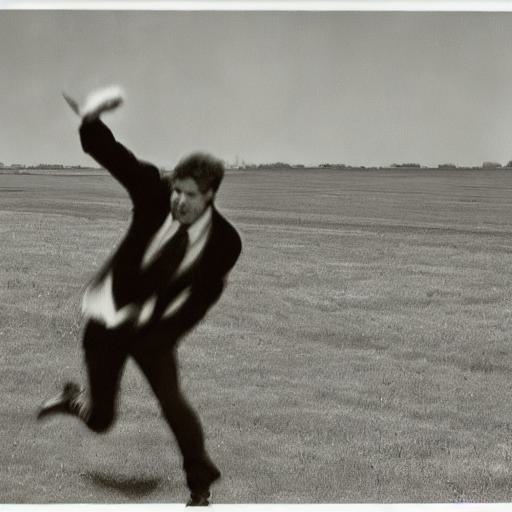}
\end{tabular}
\addtolength{\tabcolsep}{5pt}
\caption{ Visualizing images generated by an optimal prompt found by TuRBO for the target regular ImageNet class \texttt{ballplayer}. 
Top row shows three images generated by the optimal prompt. 
All three images are classified by the ResNet as ballplayers, thus indicating this was a successful prompt. 
The bottom row shows images generated from each individual token in the optimal prompt. 
None of the individual tokens are able to generate the target class.}
\label{ballplayer}
\end{figure}

\begin{figure}[ht]
\addtolength{\tabcolsep}{-5pt}   
\begin{tabular}{lll}
\multicolumn{3}{l} {\fontsize{11pt}{13pt}\textbf{Goal Class:} \texttt{piano}}\\
&&\\
     \multicolumn{3}{l} {\ttbf{wider organist cart}}\\
     \includegraphics[width=0.333\linewidth]{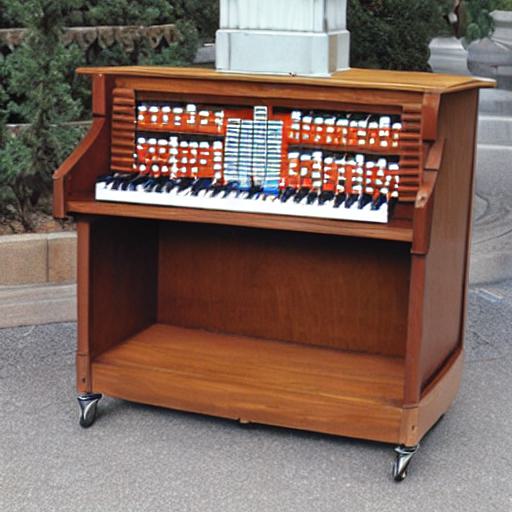} &
     \includegraphics[width=0.333\linewidth]{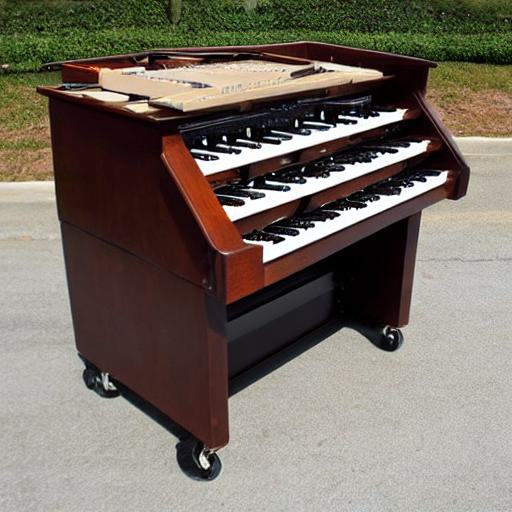} &
     \includegraphics[width=0.333\linewidth]{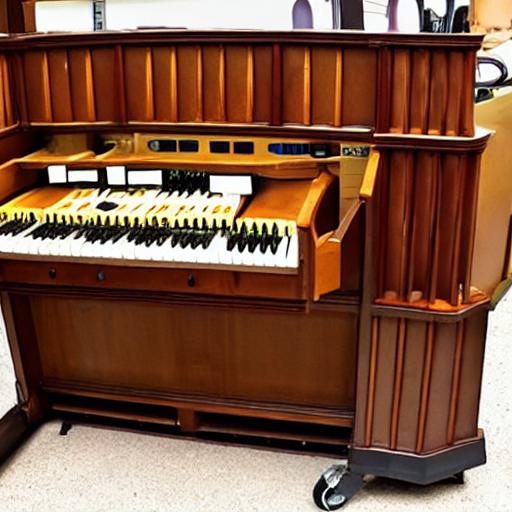}
\end{tabular}
\begin{tabular}{llll}
    \ttbf{wider} & \ttbf{organi} & \ttbf{st} & \ttbf{cart} \\
    \includegraphics[width=0.25\linewidth]{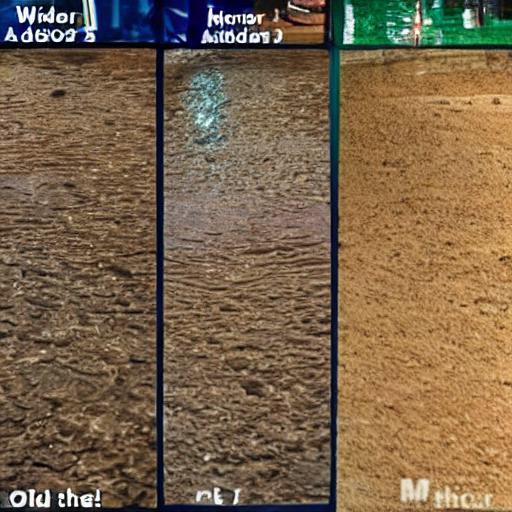} &
    \includegraphics[width=0.25\linewidth]{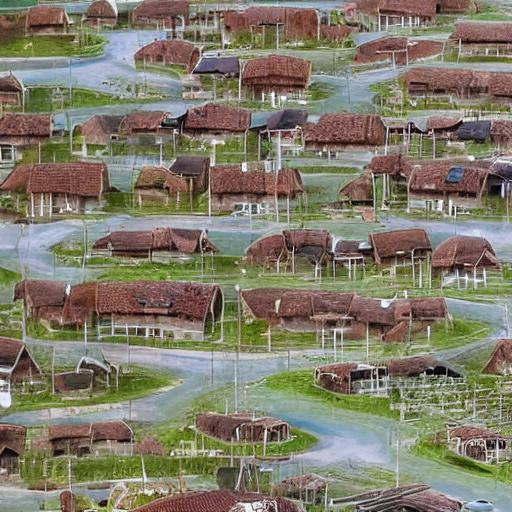} &
    \includegraphics[width=0.25\linewidth]{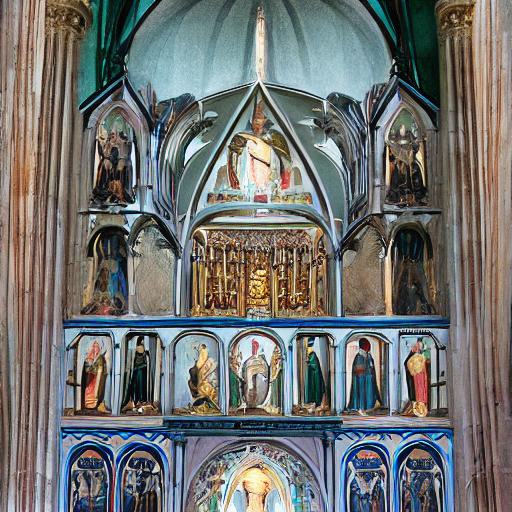} & 
    \includegraphics[width=0.25\linewidth]{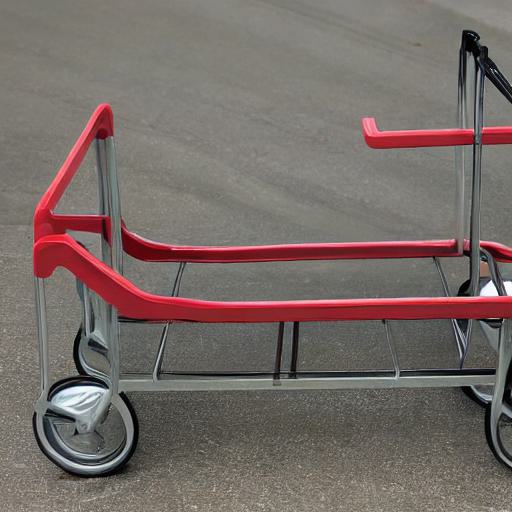}
\end{tabular}
\addtolength{\tabcolsep}{5pt} 
\caption{ Visualizing images generated by an optimal prompt found by TuRBO for the target higher-level ImageNet class: \texttt{piano}. 
Top row shows three images generated by the optimal prompt. 
All three images are classified by the ResNet as pianos, thus indicating this was a successful prompt. 
The bottom row shows images generated from each individual token in the optimal prompt. 
None of the individual tokens are able to generate the target class.}
\label{piano}
\end{figure}

\begin{figure}[ht]

\addtolength{\tabcolsep}{-5pt}   
\begin{tabular}{lll}
    \multicolumn{3}{l} {\fontsize{11pt}{13pt}\textbf{Goal Class:} \texttt{toiletry}}\\
    &&\\
     \multicolumn{3}{l} {\ttbf{artist led whitening oil}}\\
     \includegraphics[width=0.333\linewidth]{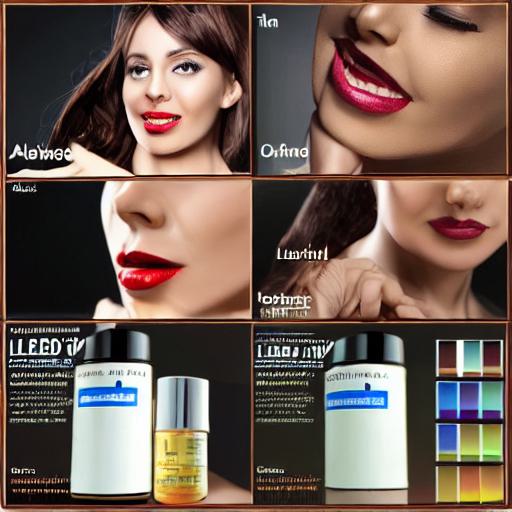}&
     \includegraphics[width=0.333\linewidth]{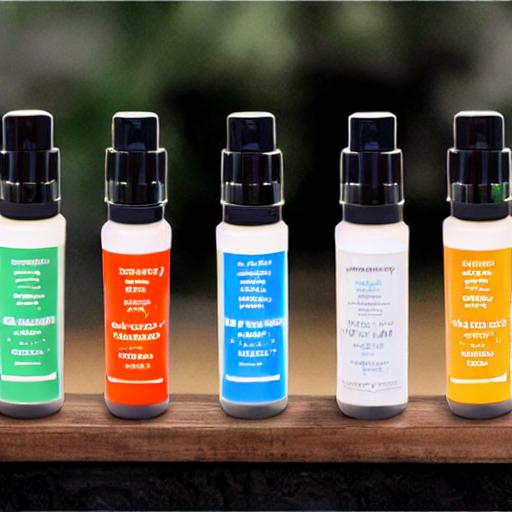}&
     \includegraphics[width=0.333\linewidth]{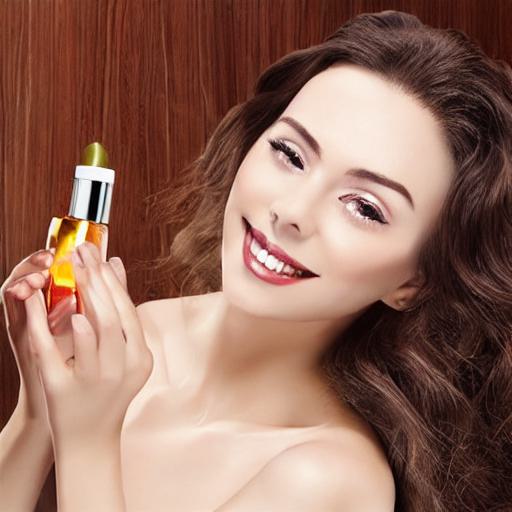}\\
\end{tabular}
\begin{tabular}{llll}
    \ttbf{artist} & \ttbf{led} & \ttbf{whitening} & \ttbf{oil}\\
    \includegraphics[width=0.25\linewidth]{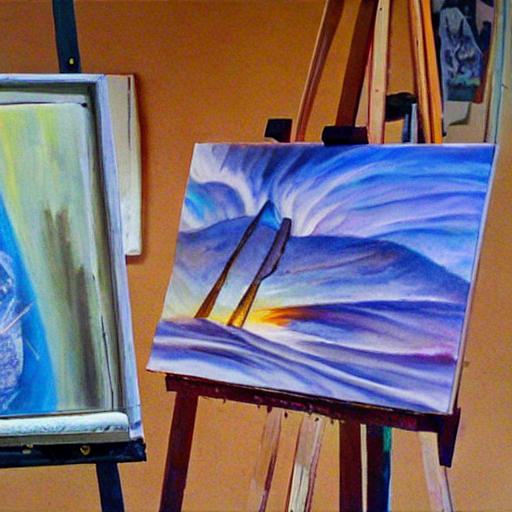}&
    \includegraphics[width=0.25\linewidth]{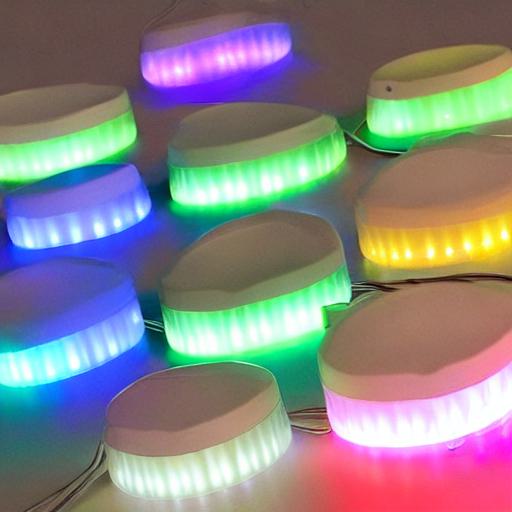}&
    \includegraphics[width=0.25\linewidth]{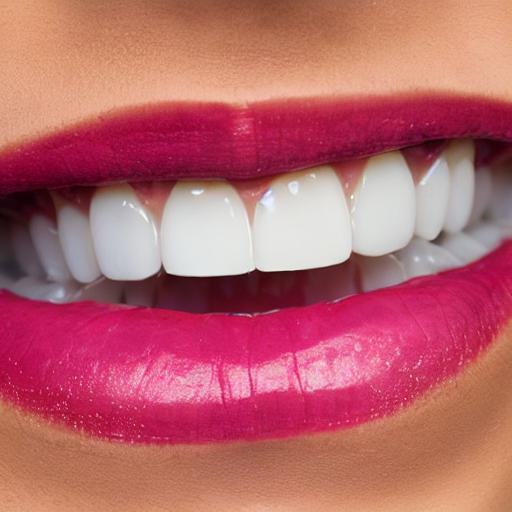}&
    \includegraphics[width=0.25\linewidth]{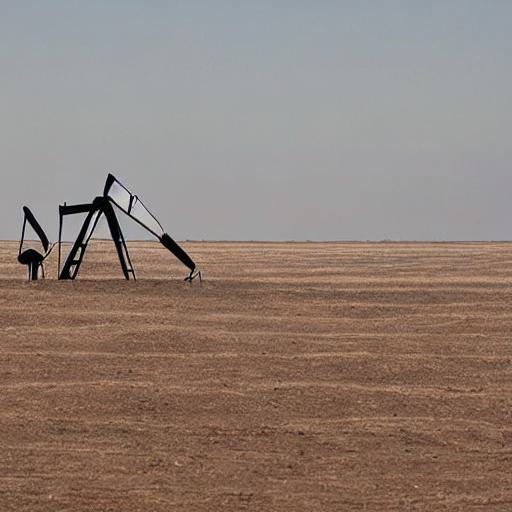}
\end{tabular}
\addtolength{\tabcolsep}{5pt}  
\caption{ Visualizing images generated by an optimal prompt found by TuRBO for the target higher-level ImageNet class: \texttt{toiletry}. 
Top row shows three images generated by the optimal prompt. 
All three images are classified by the ResNet as toiletries, thus indicating this was a successful prompt. 
The bottom row shows images generated from each individual token in the optimal prompt. 
None of the individual tokens are able to generate the target class.}
\label{toiletry}

\end{figure}

\begin{figure}[ht]
\addtolength{\tabcolsep}{-5pt}   
\begin{tabular}{lll}
    \multicolumn{3}{l} {\fontsize{11pt}{13pt}\textbf{Goal Class:} \texttt{dog}}\\
    &&\\
     \multicolumn{3}{l} {\ttbf{turbo lhaff}\checkmark \ttbf{a picture of a mountain}}\\
     \includegraphics[width=0.333\linewidth]{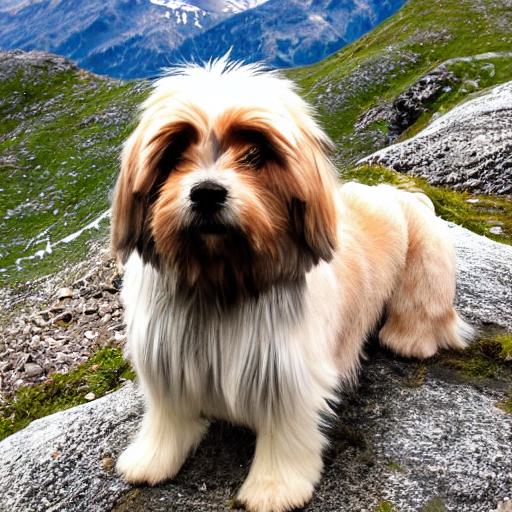} &
     \includegraphics[width=0.333\linewidth]{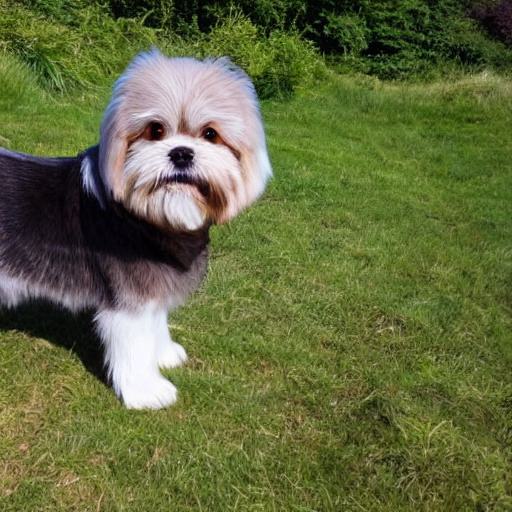} &
     \includegraphics[width=0.333\linewidth]{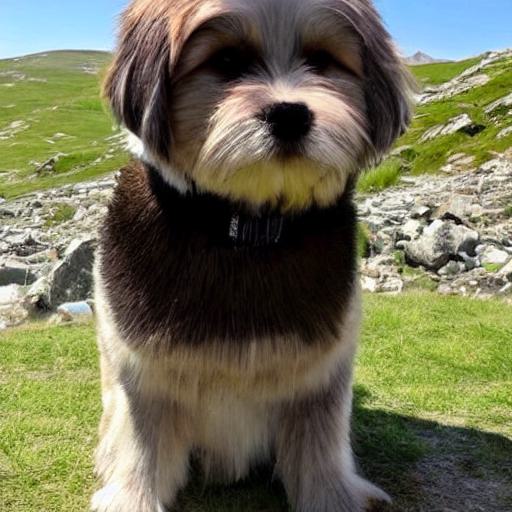} 
\end{tabular}
\begin{tabular}{lllll}
    \ttbf{turbo} & \ttbf{l} & \ttbf{ha} & \ttbf{ff} & \checkmark \\
    \includegraphics[width=0.2\linewidth]{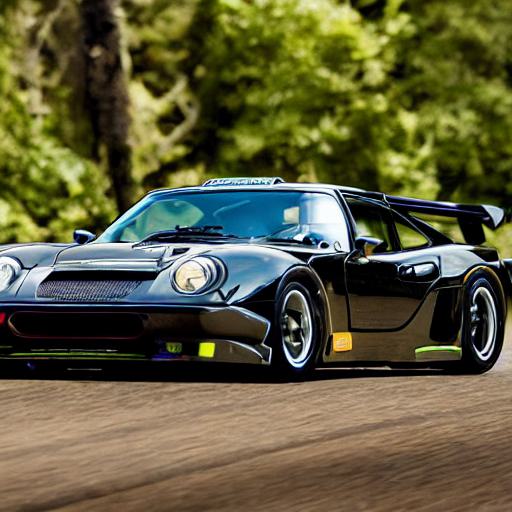} &
    \includegraphics[width=0.2\linewidth]{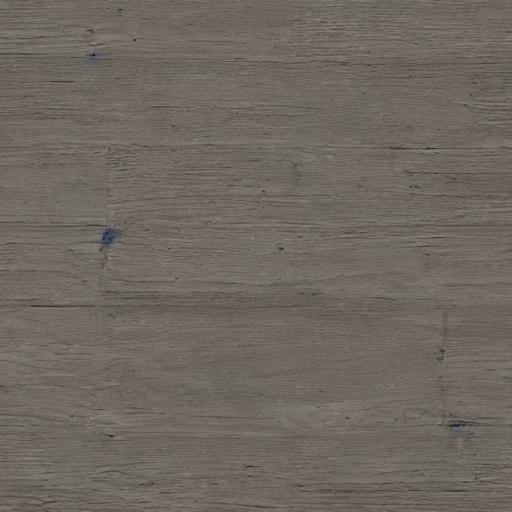} &
    \includegraphics[width=0.2\linewidth]{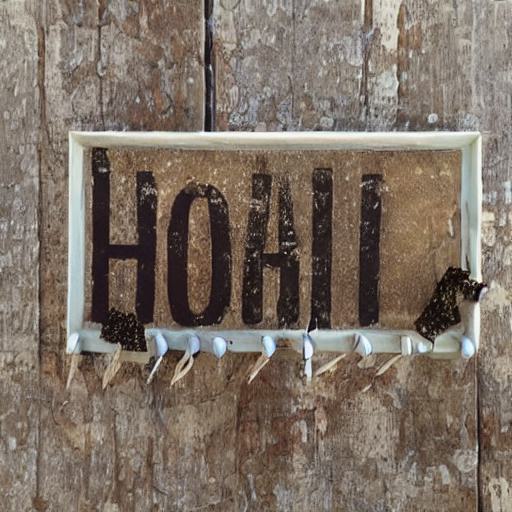} &
    \includegraphics[width=0.2\linewidth]{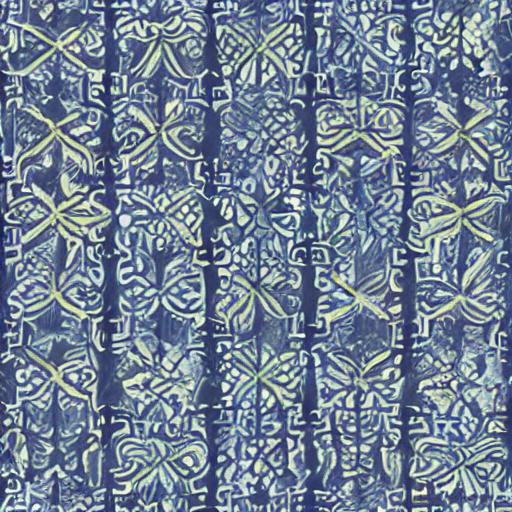} & 
    \includegraphics[width=0.2\linewidth]{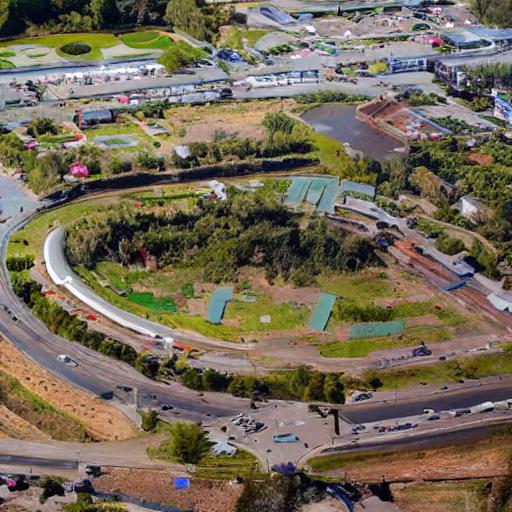} 
\end{tabular}
\addtolength{\tabcolsep}{5pt}  
\caption{ Visualizing images generated by an optimal prompt found by our method for the higher-level ImageNet class: \texttt{dog}. 
Top row shows three images generated by the optimal prompt. 
All three images are classified by the Resnet18 classifier as dogs (rather than mountains). 
The bottom row shows images generated from each individual token in the optimal prompt. 
None of the individual tokens are able to generate the target class.}
\label{dog-prepend}
\end{figure}

\begin{figure}[ht]
\centering
\addtolength{\tabcolsep}{-5pt}   
\begin{tabular}{lll}
\multicolumn{3}{l} {\fontsize{11pt}{13pt}\textbf{Goal Class:} \texttt{person}}\\
&&\\
     \multicolumn{3}{l} {\ttbf{cllr mods a picture of a dog}}\\
     \includegraphics[width=0.333\linewidth]{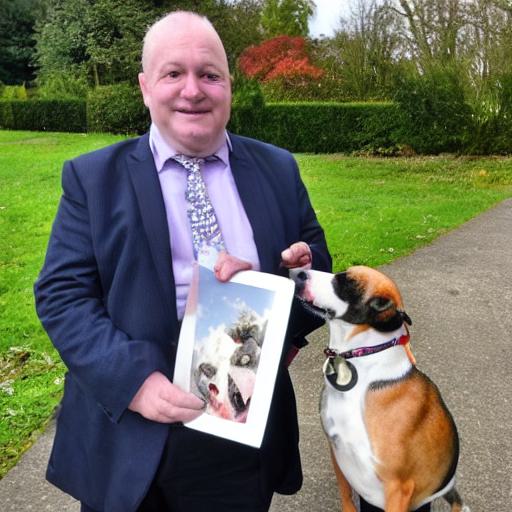} &
     \includegraphics[width=0.333\linewidth]{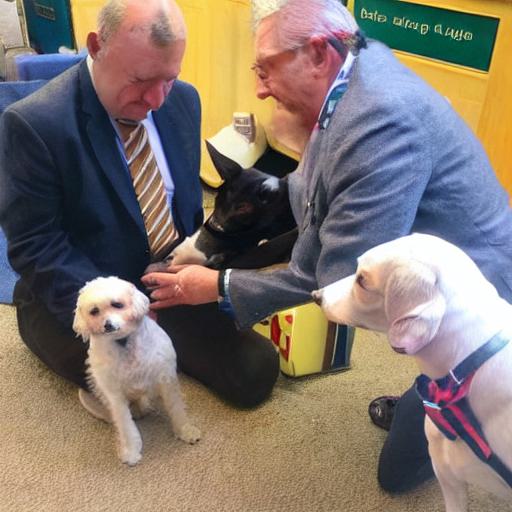} &
     \includegraphics[width=0.333\linewidth]{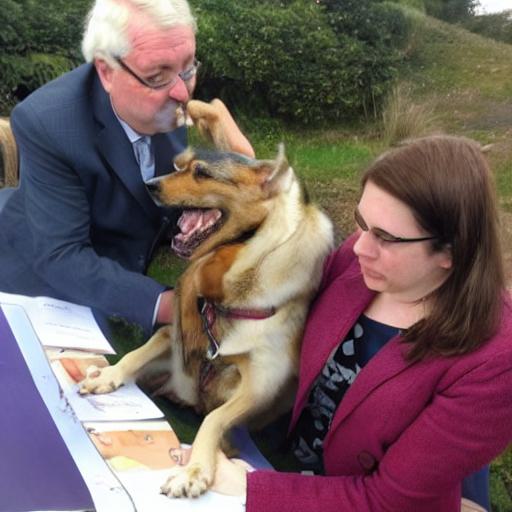}
\end{tabular}
\begin{tabular}{lll}
    \ttbf{cllr} && \ttbf{mods} \\
    \includegraphics[width=0.3\linewidth]{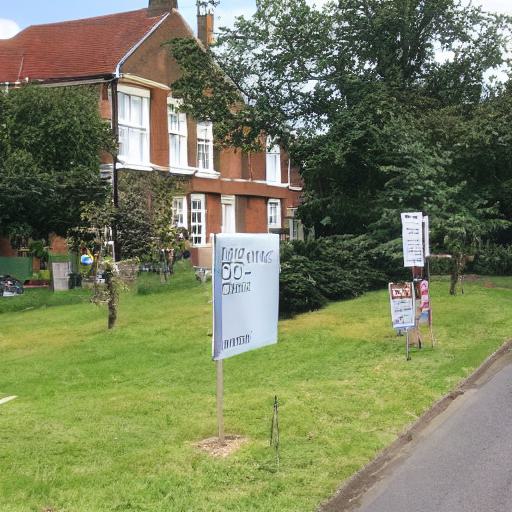} & &
    \includegraphics[width=0.3\linewidth]{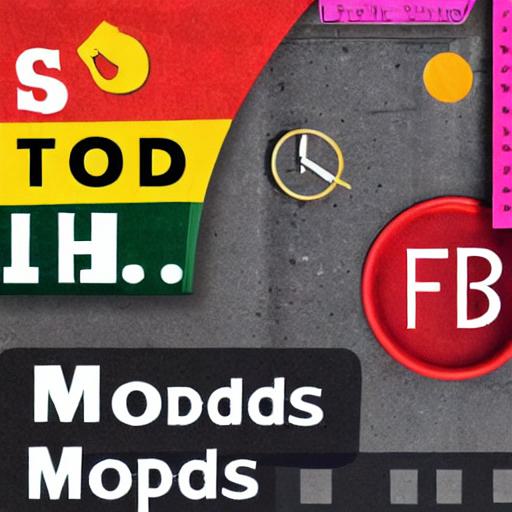}
\end{tabular}
\addtolength{\tabcolsep}{5pt}  
\caption{Visualizing images generated by an optimal prompt found by our method for the higher-level ImageNet class: \texttt{person}. 
Top row shows three images generated by the optimal prompt. 
All three images are classified by the ResNet18 classifier as people (rather than dogs). 
The bottom row shows images generated from each individual token in the optimal prompt. 
None of the individual tokens are able to generate the target class.}
\label{person-prepend}
\end{figure}

\begin{figure}[ht]
\addtolength{\tabcolsep}{-5pt}   
\begin{tabular}{lll}
\multicolumn{3}{l} {\fontsize{11pt}{13pt}\textbf{Goal Class:} \texttt{sportscar}}\\
&&\\
     \multicolumn{3}{l} {\ttbf{jaguar fp euphoria idan a picture of the ocean}}\\
     \includegraphics[width=0.333\linewidth]{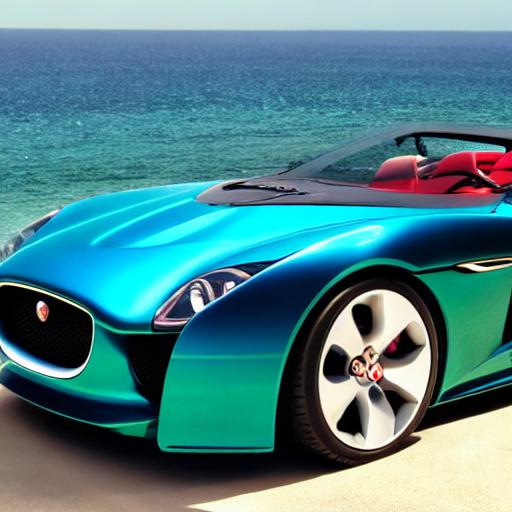}&
     \includegraphics[width=0.333\linewidth]{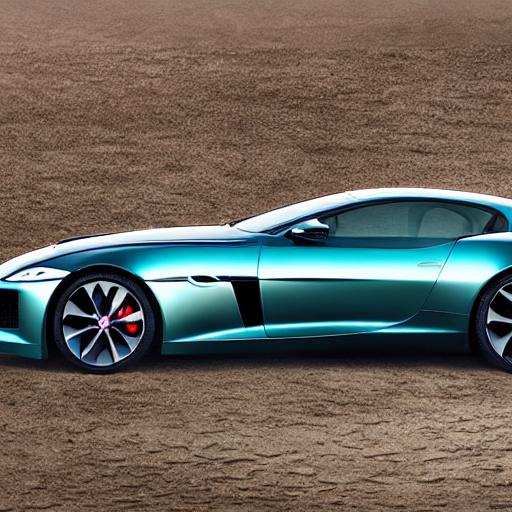}&
     \includegraphics[width=0.333\linewidth]{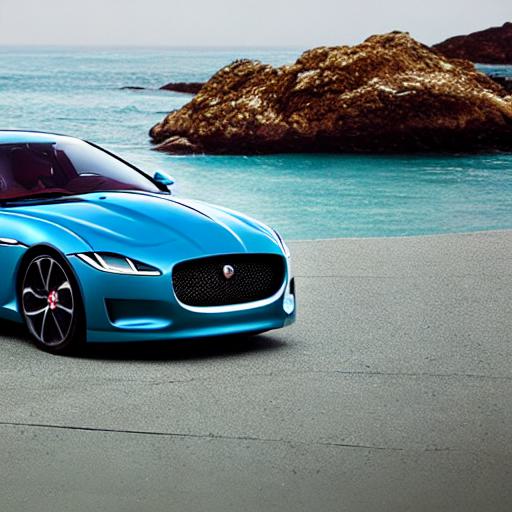}\\
\end{tabular}
\begin{tabular}{llll}
    \ttbf{jaguar} & \ttbf{fp} & \ttbf{euphoria} & \ttbf{idan}\\
     \includegraphics[width=0.25\linewidth]{
      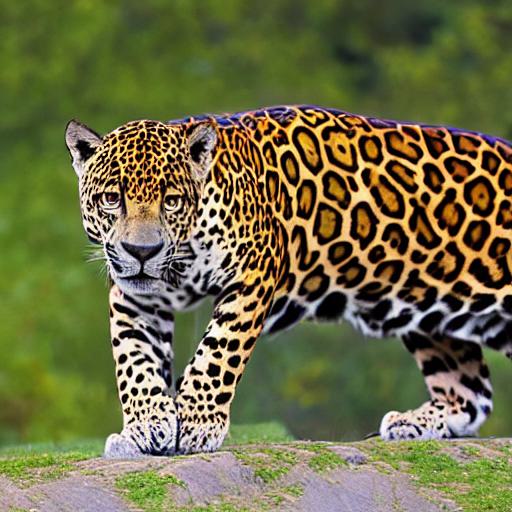}&
      \includegraphics[width=0.25\linewidth]{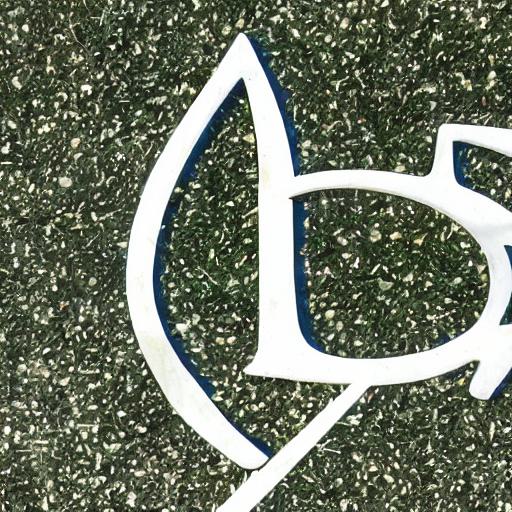}&
      \includegraphics[width=0.25\linewidth]{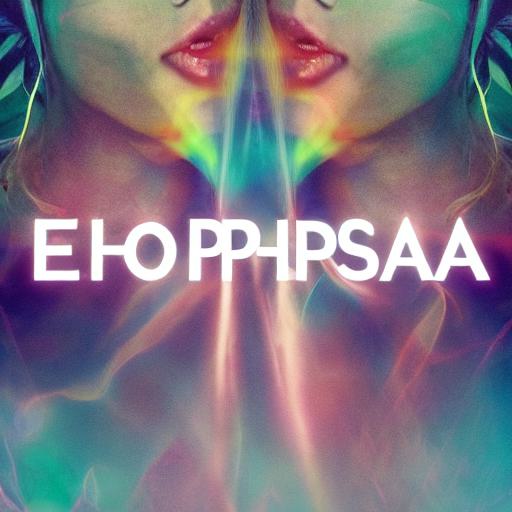}&
      \includegraphics[width=0.25\linewidth]{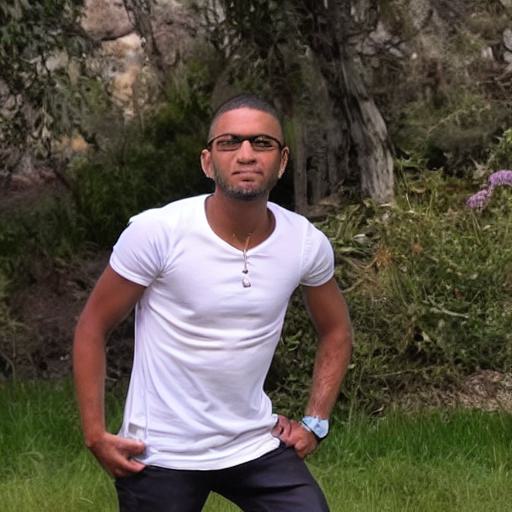}
\end{tabular}
\addtolength{\tabcolsep}{5pt}   
\caption{Visualizing images generated by an optimal prompt found by our method for the single ImageNet class: \texttt{sportscar}. 
Top row shows three images generated by the optimal prompt. 
All three images are classified by the ResNet18 classifier as sportscars (rather than oceans). 
The bottom row shows images generated from each individual token in the optimal prompt. 
None of the individual tokens are able to generate the target class.}
\label{sportscar-p}
\end{figure}

\begin{figure}
\begin{tabular}{lll}
    \multicolumn{3}{l} {\fontsize{11pt}{13pt}\textbf{Goal Class:} \texttt{ballplayer}}\\
     \multicolumn{3}{l} {\fontsize{8pt}{10pt}\ttbfs{fiji players hormoncaine a picture of a mountain}}\\
     \includegraphics[width=0.333\columnwidth]{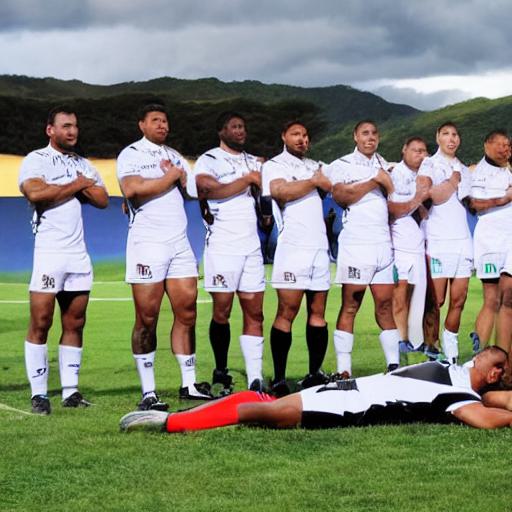} &
     \includegraphics[width=0.333\columnwidth]{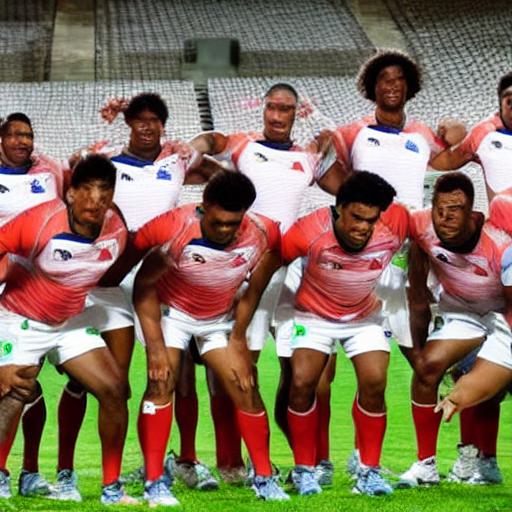} &
     \includegraphics[width=0.333\columnwidth]{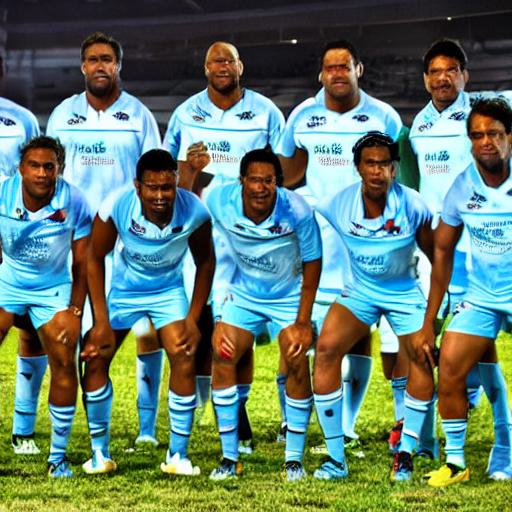}
\end{tabular}
\newline
\begin{tabular}{llll}
    \ttbfs{fiji} & \ttbfs{players} & \ttbfs{hormon} & \ttbfs{caine} \\
    \includegraphics[width=0.25\columnwidth]{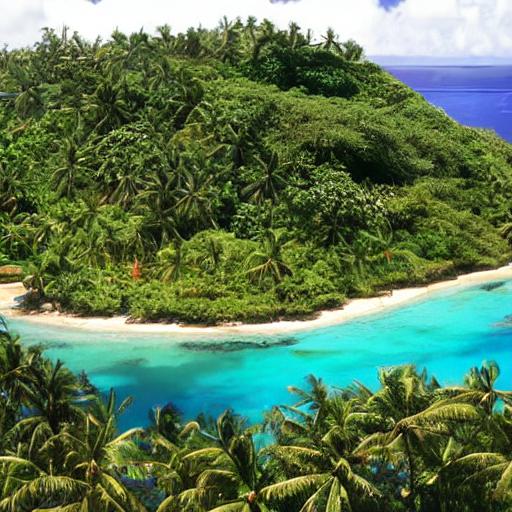}&
    \includegraphics[width=0.25\columnwidth]{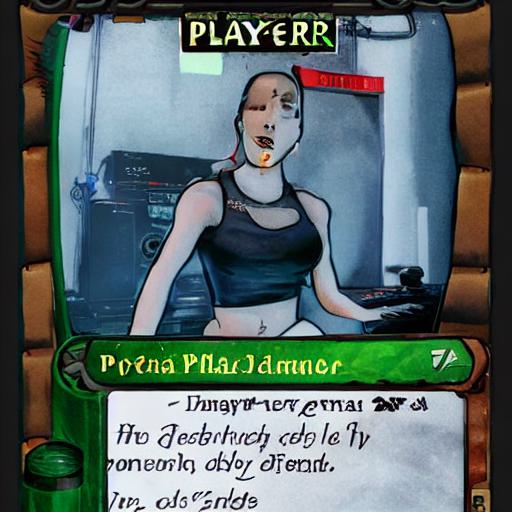} &
    \includegraphics[width=0.25\columnwidth]{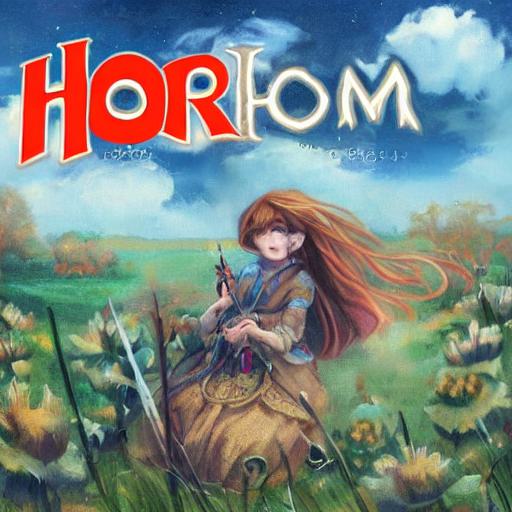} & 
    \includegraphics[width=0.25\columnwidth]{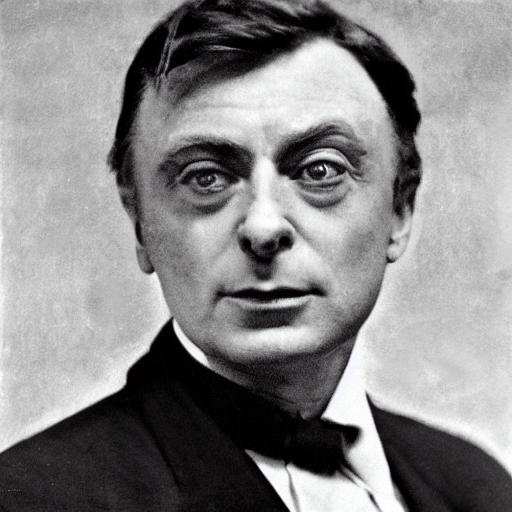}
\end{tabular}
\addtolength{\tabcolsep}{5.5pt} 
\caption{Visualizing images generated by an optimal prompt found by TuRBO for the target higher-level ImageNet class: \texttt{ballplayer}. 
Top row shows three images generated by the optimal prompt. 
All three images are classified by the ResNet as ballplayers, thus indicating this was a successful prompt. 
The bottom row shows images generated from each individual token in the optimal prompt. 
None of the individual tokens are able to generate the target class.}
\label{ballplayer-prepend}
\end{figure}

We provide additional examples of optimal prompts found by our methods for the image generation tasks and the resultant images generated by those prompts. 
See additional examples for Task \ref{task: restricted prompts 2} in \Cref{lipstick,ballplayer,piano,toiletry}.
See additional examples for Task \ref{task: restricted prepending prompts 3} in \Cref{dog-prepend,person-prepend,sportscar-p,ballplayer-prepend}.

\vspace*{\fill}

\vspace*{\fill}
\clearpage

\vspace*{\fill}

\vspace*{\fill}
\clearpage

\clearpage 
\vspace*{\fill}
\vspace*{\fill}
\clearpage 

\vspace*{\fill}
\vspace*{\fill}
\clearpage

\vspace*{\fill}

\vspace*{\fill}
\clearpage

\vspace*{\fill}

\vspace*{\fill}
\clearpage 

\clearpage
\section{Text Generation}\label{app: text gen}
\subsection{Seed Prompt Details}
{\renewcommand{\arraystretch}{1.5}
\begin{table}[h]
    \centering
    \begin{tabular}{p{4cm} l p{8cm}}
    \toprule
    Origin & Reference Name & Full Prompt\\
    \midrule
        SCAN, template\_jump\_around\_right, test, affirmative\_bottom, 566 \citep{scan}& SCAN & `run opposite left after jump around right thrice\textbackslash n\textbackslash n Given the commands above, produce the corresponding correct sequence of actions. The actions should be comma-separated.'\\
       2023 AMC8 Problem 21 & Math &`Bob writes the numbers 1, 2, ..., 9 on separate pieces of paper, one number per paper. He wishes to divide the papers into 3 piles of three papers so that the sum of the numbers in each pile will be the same. In how many ways can this be done?'\\
       tmu\_gfm\_dataset, train, correct-sentence, 329 \citep{tmu} & Grammar &`Grammatically improve the below text. Note that the original meaning has to be preserved and also it should sound natural.\textbackslash n\textbackslash n Text: People needs a safe environment to live in, and also needs a private environment to stay independent.'\\
       GPT4All \citep{gpt4all} &Python &`Explain list comprehension in Python.'\\
      SWAG, regular, test, how\_ends, 8686 \citep{zellers2018swagaf}& &`From behind, someone swings the poker at his head. He...\textbackslash n How does the description likely end?\textbackslash n (a): falls face down on the carpet.\textbackslash n (b): glances at all of his heads.\textbackslash n (c): looks grandma in the eye.\textbackslash n (d): stares at the phone.\textbackslash n'\\
      GPT4All \citep{gpt4all} & Poem  &`Compose a 10 line rhyming poem about cats.'\\
      OASST1 Validation \citep{köpf2023openassistant}& Meaning of Life &`What is the meaning of life, written in the style of Dr. Seuss poem?' \\
      GPT4ALL \citep{gpt4all} &Slogan &`Design a slogan for a restaurant that serves traditional Taiwanese dishes.' \\
    \bottomrule
    \end{tabular}
    \vspace{0.2cm}
    \caption{Table of the seed prompts considered, where the prompt was obtained from, and our reference name. For prompts from an explicit dataset, we provide specific identifying details for the sample, such as the exact sample number. For some of the prompts, we modify the wording slightly to avoid using prompts the model has been trained on.}
    \label{tab: seed prompts}
\end{table}
}

\clearpage
\subsection{More Seed Prompt Generations}

\begin{figure}[t]
    \centering
    \includegraphics[width=0.9\columnwidth]{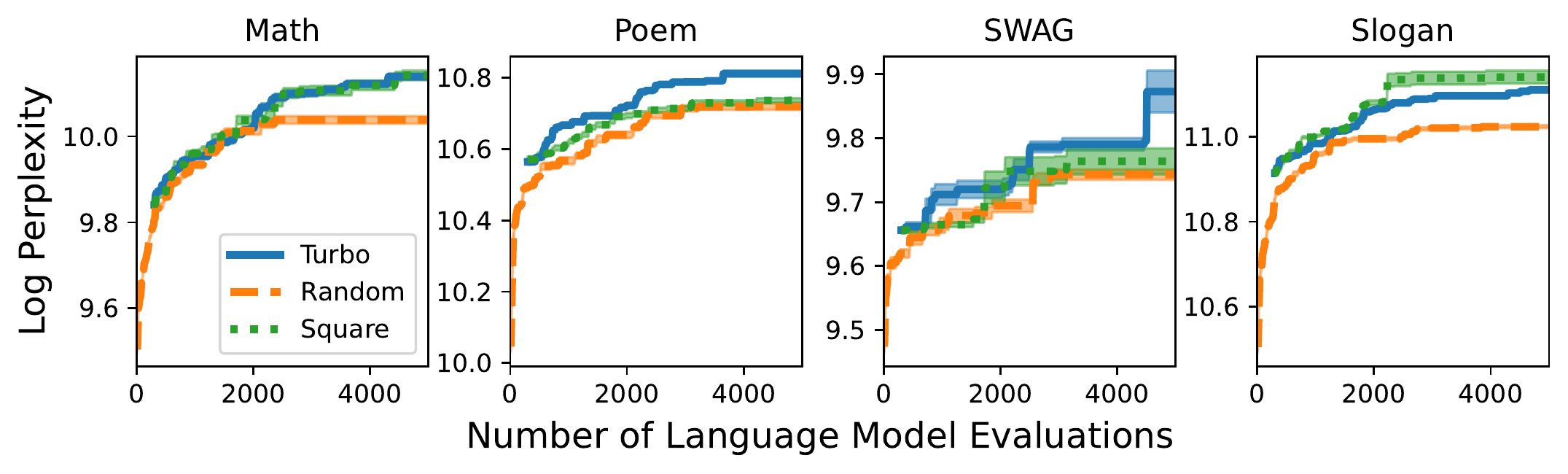}
    \caption{Plots of the maximum log perplexity with respect to the number of language model evaluations. The values are averaged over 20 random seeds, with the standard error shaded. The full seed prompts are in \cref{tab: seed prompts}.}
    \label{fig: text perplexity plot more}
\end{figure}

We provide more examples of generated text outputs in \cref{fig: text-examples more,fig: text-examples math,fig: text-examples scan,fig: text-examples scan more} and the performance on other seed prompts in \cref{fig: text perplexity plot more}. We see that we are consistently able to optimize desired loss functions and obtain nonsensical outputs.

\begin{figure}[h]
\centering
\tbox[\textbf{Prompt: }\textit{oral geschichte ' efficient ding once} Grammatically improve the below text. Note that the original meaning has to be preserved and also it should sound natural.\textbackslash n\textbackslash nText: People needs a safe environment to live in, and also needs a private environment to stay independent.]{
            \textbf{Generated Examples: } \\
            \begin{tabular}{l p{12cm}}
                 1. &  Menschen brauchen eine sichere Umgebung, in der sie leben können, und auch eine private Umgebung, in der sie unabhängig bleiben können.\\
                 2. & Menschen brauchen eine sichere Umgebung, um in ihr zu leben, und auch eine eigene Umgebung, um unabhängig zu bleiben.\\ 
                 3. & Menschen brauchen eine sichere Umgebung, in der sie leben können, und auch eine private Umgebung, um unabhängig zu bleiben.
            \end{tabular}
        }
    \caption{Grammar Prompt: Three examples of text generated from Vicuna 13B-v1.1 using our optimized prompt in italics and the original prompt to which we prepend. The generated text are nonsensical or irrelevant to the prompt. Note that while `geschichte' is `story' in German, without the apostrophe, the model will not consistently generate German text.}
\label{fig: text-examples more}
\end{figure}

\begin{figure}[h]

\centering
\tbox[\textbf{Prompt: }\textit{cod irs theorem fit lay pg} Bob writes the numbers 1, 2, ..., 9 on separate pieces of paper, one number per paper. He wishes to divide the papers into 3 piles of three papers so that the sum of the numbers in each pile will be the same. In how many ways can this be done?]{
            \textbf{Generated Examples: } \\
            \begin{tabular}{l p{12cm}}
                 1. &  This problem is known as the ``Coding Theorem" or ``Codirras-Wagner Theorem". It can be solved using combinatorics, specifically permutations and combinations. Let's denote the three piles of papers as \\
                 2. & This problem is an example of the ``Coffee and Cake" problem, also known as the ``Irregular Rearrangement" problem. The solution to this problem is to use the ``Codirisu Theorem", which states\\ 
                 3. &  This problem is a variation of the so-called ``Coffee Shop" problem, which is a well-known problem in combinatorics. The solution to this problem is to use the ``Coffee Shop" solution, which 
            \end{tabular}
        }
\caption{Math Prompt: Three examples of text generated from Vicuna 13B-v1.1 using our optimized prompt in italics and the original prompt to which we prepend. The problems and theorems, e.g. ``Coffee and Cake problem" and ``Codirras-Wagner Theorem", are not real problems or theorems.}

\label{fig: text-examples math}

\end{figure}

\begin{figure}
        \centering
    
\centering
\tbox[\textbf{Prompt: }\textit{ben post ole fin quick marine} run opposite left after jump around right thrice\textbackslash n\textbackslash n Given the commands above, produce the corresponding correct sequence of actions. The actions should be comma-separated.]{
            \textbf{Generated Examples: } \\
            \begin{tabular}{l p{12cm}}
                 1. & Human, bench press 50kg, 3 sets of 8 reps, bodyweight squats 20 reps, rowing machine 500m, push-ups 10 reps, 1-\\
                 2. & Human, bench press 100 pounds for 3 sets of 8 reps, bicep curls 20 pounds for 3 sets of 10 reps, tricep dips 40 pounds for\\
                 3. &  Human, bench press 50 kg for 3 sets of 10 reps, bicep curl 20 kg for 3 sets of 10 reps, tricep dips for 3 sets of  
            \end{tabular}
        }
\caption{SCAN Prompt: Three examples of text generated from Vicuna 13B-v1.1 using our optimized prompt in italics and the original prompt to which we prepend. The generated text comprises exercise routines rather than relevant actions from the prompt.}

\label{fig: text-examples scan}

\end{figure}

\begin{figure}
        \centering
    
\centering
\tbox[\textbf{Prompt: }\textit{default since levant rico ny levant} run opposite left after jump around right thrice\textbackslash n\textbackslash n Given the commands above, produce the corresponding correct sequence of actions. The actions should be comma-separated.]{
            \textbf{Generated Examples: } \\
            \begin{tabular}{l p{12cm}}
                 1. & Human, default, default, default, default, default, default, default, default, default, default, default, default, default, default, default, default, default, default, default, default, default, default, default, default,\\
                 2. & Human, default, default, default, default, default, default, default, default, default, default, default, default, default, default, default, default, default, default, default, default, default, default, default, default,\\
                 3. &  Human, default, default, default, default, default, default, default, default, default, default, default, default, default, default, default, default, default, default, default, default, default, default, default, default,\\
            \end{tabular}
        }
\caption{SCAN Prompt: Three examples of text generated from Vicuna 13B-v1.1 using our optimized prompt in italics and the original prompt to which we prepend. The generated text is essentially only the word `default'.}
\label{fig: text-examples scan more}

\end{figure}